\documentclass[journal]{IEEEtran}
\usepackage{amsmath,amsfonts}
\usepackage{amssymb}
\usepackage{algorithmic}
\usepackage{algorithm}
\usepackage{array}
\usepackage{capt-of}
\usepackage[caption=false,font=footnotesize]{subfig}
\usepackage{booktabs}
\usepackage{textcomp}
\usepackage{stfloats}
\usepackage{verbatim}
\usepackage{graphicx}
\usepackage{cite}
\usepackage{multirow}
\usepackage{xspace}
\usepackage[hyphens]{url}
\usepackage{breakurl}
\usepackage[breaklinks=true,bookmarks=false]{hyperref}
\usepackage{orcidlink}
\usepackage{makecell}
\usepackage{pifont}
\newcolumntype{L}[1]{>{\raggedright\arraybackslash}m{#1}}
\newcolumntype{C}[1]{>{\centering\arraybackslash}m{#1}}
\newcolumntype{R}[1]{>{\raggedleft\arraybackslash}m{#1}}
\newcolumntype{P}[1]{>{\centering\arraybackslash}p{#1}}
\usepackage[capitalize]{cleveref}
\crefname{section}{Sec.}{Secs.}
\Crefname{section}{Section}{Sections}
\Crefname{table}{Table}{Tables}
\crefname{table}{Tab.}{Tabs.}
\makeatletter
\DeclareRobustCommand\onedot{\futurelet\@let@token\@onedot}
\def\@onedot{\ifx\@let@token.\else.\null\fi\xspace}

\def\eg{\emph{e.g}\onedot} 
\def\ie{\emph{i.e}\onedot} 
 
\def\etc{\emph{etc}\onedot} \def\vs{\emph{vs}\onedot}
 
\def\etal{\emph{et al}\onedot}
\makeatother
\begin{document}

\title{PGNeXt: High-Resolution Salient Object Detection \\via Pyramid Grafting Network}

\author{Changqun Xia, Chenxi Xie, Zhentao He, Tianshu Yu, and Jia Li,~\IEEEmembership{Senior Member,~IEEE}
        
\thanks{This work was supported by the National Natural Science Foundation of China under the Grant 62132002 and Grant 62102206.}
\thanks{Changqun Xia and Chenxi Xie contribute equally to this work.}
\thanks{Changqun Xia is with the Peng Cheng Laboratory, Shenzhen\,518055, China.}
\thanks{Chenxi Xie, Zhentao He, Tianshu Yu and Jia Li are with the State Key Laboratory of Virtual Reality Technology and Systems, School of Computer Science and Engineering, Beihang University, Beijing 100191, China.}
\thanks{Jia Li is the corresponding author (E-mail: \href{mailto:jiali@buaa.edu.cn}{jiali@buaa.edu.cn}).}
\thanks{A preliminary version of this work has appeared in CVPR 2022~\cite{PGNet}.}
}

\markboth{Journal of \LaTeX\ Class Files,~Vol.~14, No.~8, August~2021}%
{Shell \MakeLowercase{\textit{et al.}}: A Sample Article Using IEEEtran.cls for IEEE Journals}

\IEEEpubid{0000--0000/00\$00.00~\copyright~2021 IEEE}

\maketitle

\begin{abstract}
We present an advanced study on more challenging high-resolution salient object detection (HRSOD) from both dataset and network framework perspectives. To compensate for the lack of HRSOD dataset, we thoughtfully collect a large-scale high resolution salient object detection dataset, called UHRSD, containing 5,920 images from real-world complex scenarios at 4K-8K resolutions. All the images are finely annotated in pixel-level, far exceeding previous low-resolution SOD datasets. Aiming at overcoming the contradiction between the sampling depth and the receptive field size in the past methods, we propose a novel one-stage framework for HR-SOD task using  pyramid grafting mechanism. In general, transformer-based and CNN-based backbones are adopted to extract features from different resolution images independently and then these features are grafted from transformer branch to CNN branch. An attention-based Cross-Model Grafting Module (CMGM) is proposed to enable CNN branch to combine broken detailed information more holistically, guided by different source feature during decoding process. Moreover, we design an Attention Guided Loss (AGL) to explicitly supervise the attention matrix generated by CMGM to help the network better interact with the attention from different branches. Comprehensive experiments on UHRSD and widely-used SOD datasets demonstrate that our method can simultaneously locate salient object and preserve rich details, outperforming state-of-the-art methods. To verify the generalization ability of the proposed framework, we apply it to the camouflaged object detection (COD) task. Notably, our method performs superior to most state-of-the-art COD methods without bells and whistles.

\end{abstract}


\section{Introduction}

\IEEEPARstart{S}{alient} object detection (SOD) aims to locate the most attractive objects in the scene and accurately segment their structures. It has been used as a pre-processing step that can be widely applied to a variety of computer vision tasks, such as light field segmentation\cite{liu2021light,zhang2019memory}, instance segmentation \cite{zhou2020multi}, video object segmentation\cite{ji2021full,zhang2021dynamic} and \etc \cite{ge2017detecting,ge2018low,zhao2021graph,he2019part}. Benefiting from the advances in deep learning \cite{ImageNet,li2022selective,qiao2019transductive,chen2020learning}, SOD methods are no longer limited to the hand-crafted features, leading to robustness in various complex scenarios. Most of SOD methods employ the feature-pyramid-network (FPN)-style framework utilizing multi-level features that form a bottom-up decoding structure gradually recovering details of the saliency maps, and they have achieved remarkable achievements\cite{chen2018reverse,su2019banet,liu2020dynamic,qin2020u2,ji2021calibrated,fan2020bbs}. Although the existing methods have achieved impressive performance on several low-resolution (LR) benchmark datasets, there are limited attention given to high-resolution (HR) SOD task.
\begin{figure}[t]
    \centering
    \includegraphics[width=\linewidth]{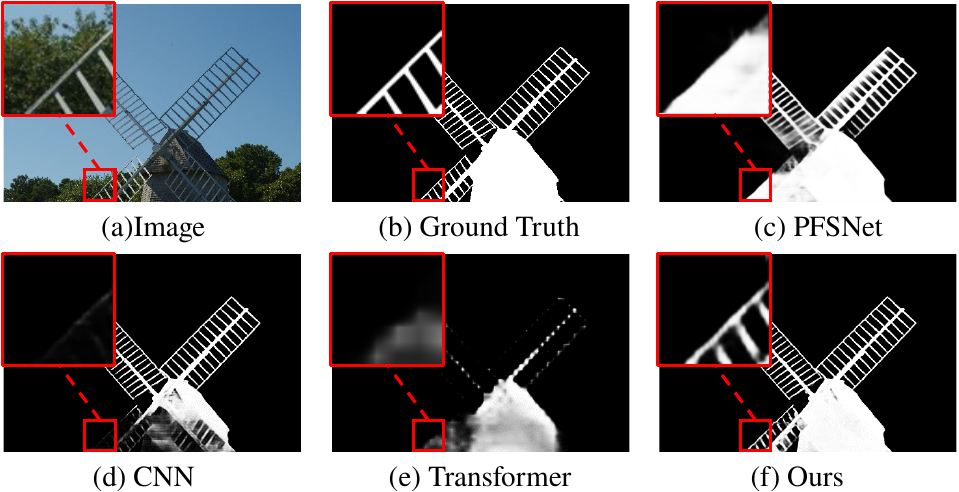}
    \caption{\textbf{Comparison of the results of the different methods.} (a) Input image. (b) Ground truth maks. (c) Down-sample then input to CNN-based PFSNet.(d) Directly input to ResNet-18 based FPN. (e) Down-sample then input to Swin-based FPN. (f) Ours.}
    \label{fig:first}
\end{figure}

\IEEEpubidadjcol

As daily accessible images are stepping towards high-resolution (\eg 1080p, 2K and 4K), the ability to process high-resolution inputs is becoming increasingly important for SOD methods. Therefore, the advanced SOD models should be able to deal with the HR inputs to better cope with the requirements of real-world applications. However, existing SOD methods are mostly designed on the datasets at low-resolution (lower than $512 \times 512$ pixels), and applying them directly to high-resolution inputs leads to a series of issues.
In \cref{fig:first}, we illustrate several typical failures of several existing SOD methods in processing HR images. \cref{fig:first} (c) and \cref{fig:first} (e) show a straightforward scheme where the image is down-sampled to a regular resolution and fed into the networks to obtain the saliency maps, and it is evident that although the overall region of the salient object is identified the fine details are severely corrupted. Besides, if we adopt high-resolution inputs without downsampling, it may lead to an incomplete structure of the salient object, as shown in \cref{fig:first} (d). Most of the existing LR SOD methods cannot simultaneously meet the requirements of ensuring that the salient object is accurate and complete while retaining sufficient details and sharp edges with high-resolution input. Thus we can observe the challenge of HR-SOD task, which is to simultaneously balance the global integrity of the images as well as local details for high-resolution images. Besides, the inference speed is also an important requirement while ensuring accuracy.

\begin{figure}[t]
    \centering
    \includegraphics[width=\linewidth]{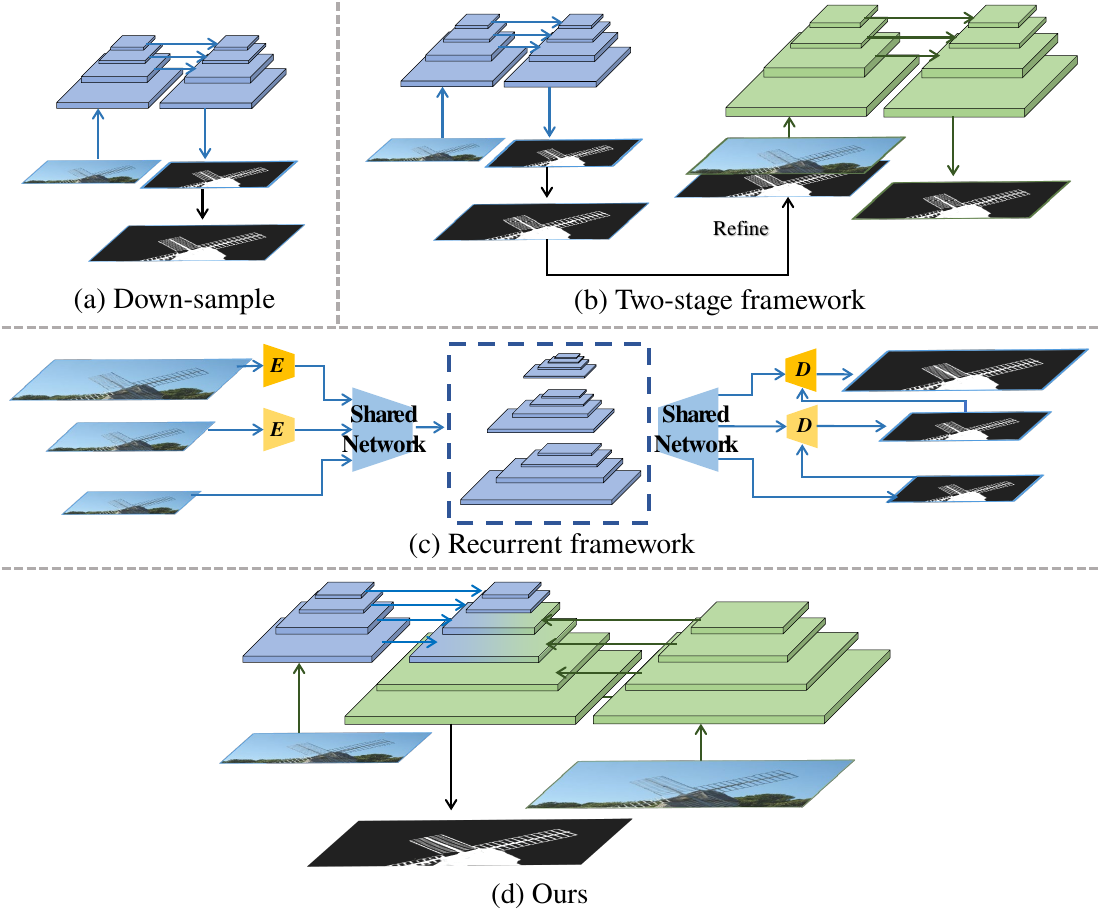}
    \caption{\textbf{Comparison of different methods for HR inputs.} (a) Down-sampling strategy. (b) Two-stage coarse to fine framework. (c) Recurrent framework (d) Pyramid grafting framework.}
    \label{fig:2}
\end{figure}

To address the above challenges, researchers have begun to customize the strategies for the high-resolution input. 
The most straightforward scheme is shown in \cref{fig:2} (a), which is to first down-sample the images to obtain the saliency maps in low-resolution and then up-sample the coarse maps as the final results. In this way, many fine details of the image are inevitably lost during the downsampling process, and the size of extracted feature maps is small, which makes it difficult to recover the original fine structure and edge details in the bottom-up decoding process.
Subsequently, to improve the process of simple up-sampling of the coarse maps, two-stage coarse-to-fine framework has been proposed \cite{zeng2019towards}, \cite{tang2021disentangled}. As shown in \cref{fig:2} (b), the first network captures global semantics at low-resolution to locate salient objects and generate a rough saliency map. The rough prediction map is then fed at high-resolution to another local network for refinement, recovering the detailed structure and share edges. However, since some small objects are ignored on the low-resolution intermediate map and the high-resolution refining process is based on this map, these objects cannot be recovered at the high-resolution stage. Therefore, such a multi-stage design leads to inconsistent contextual semantic transfer between stages. 
Lately, a new recurrent framework is proposed \cite{RMFormer} that avoids the drawbacks associated with multi-stage framework, which is shown in \cref{fig:2} (c). The key of this framework is to use the same network to process inputs from low-resolution to high-resolution in a recurrent way and refine the multi-scale intermediate results. Although this strategy shows better performance over previous ones, the multiple inferences lead to a drastic increase in computation burden.

By analyzing the shortcomings of the strategies, we can find that the key issue is that the specific features in a solitary network cannot settle the paradox of receptive field and detail retention simultaneously. In other words, traditional FPN-based networks can only extract a set of features with a limited range of resolution variations, which is impossible to cover both the required global semantics and rich details. Taking into above analysis, there are two main principles should be considered for the framework design. 

1) Considering that limited range of feature resolution fails to settle the paradox between context and details, features with a larger range of resolution should be preserved so that a higher feature pyramid can be constructed to recover rich details.

2) Considering the fact that computation grows with the square of feature resolution, high and low resolution features should be treated unequally, balancing the problem of computational burden caused by high-resolution input, thus improving the overall efficiency of the network.

Regarding the above design criteria, we can asymmetrically extract two or more sets of features of different spatial sizes and then transfer the information from one branch to the other in the same network. As illustrated in \cref{fig:2} (d), we construct a higher feature pyramid fully preserving high-resolution details and global semantics, and it can be trained end-to-end to avoid inconsistent semantics. Based on this architecture, we propose a novel one-stage deep neural network for high-resolution saliency detection named Pyramid Grafting Network (PGNet). Specifically, we use both ResNet and Transformer as our encoders, extracting features with dual spatial sizes in parallel. The transformer branch first decode the features in the FPN style, then pass the global semantic information to the ResNet branch in the stage where the feature maps of two branches have similar spatial sizes. Eventually, the ResNet branch completes the decoding process with the grafted features. Compared to classic FPNs, we have constructed a higher feature pyramid at a lower cost. We call this process feature grafting. To better graft heterologous features cross two different types of models, we design the Cross-Model Grafting Module (CMGM) based on the attention mechanism and propose the Attention Guided Loss (AGL) to further guide the grafting.
With CMGM and AGL, the heterologous features between multiple branches in parallel interact with each other to form complementary effects, eventually forming the grafting network.
Considering that supervised deep learning method requires a large amount of high quality data, we have provided a 4K resolution SOD dataset (UHRSD) with the largest number to date in an effort to promote future high-resolution salient object detection research.

This paper is an extended version of our previous work\cite{PGNet}. In particular, (a) we provide more statistics and analysis on the proposed UHRSD dataset. (b) Based on the pyramid grafting mechanism, we further explored and found that progressively staggered stacking of branches can be applied to higher resolution input and achieve better performance. As an upgrade to the conference version, we distinguish the improved version based on hierarchical pyramid grafting framework as PGNeXt. (c) We take the window-based cross attention to replace the original global cross attention in CMGM, achieving improved grafting performance while significantly reducing computational overhead, and this modified version is referred to as wCMGM. (d) We conduct more thorough experiments and ablation studies to validate the effectiveness of the network  and each module within it. (e) We partition subsets for UHRSD based on the complexity of salient objects and provide benchmark on them for HR-SOD task. (f) We apply our method to the highly associated camouflaged object detection (COD) task to demonstrate the generalization ability of our proposed pyramid grafting strategy.

The following sections of the article are organized as follows. Sec II. briefly reviews previous work related to this work. Sec III. provides more details and analysis on proposed UHRSD dataset. Sec IV. thoroughly describes our new framework of PGNeXt and improvements of window-based Cross-Model Grafting Module.

Sec V. provides SOD methods for benchmarking on the existing HR-SOD dataset and our UHRSD dataset to facilitate a better understanding of the differences between LR-SOD and HR-SOD tasks. Besides, it also gives and more ablation study of each module. Sec VI. presents the experimental results on COD tasks to demonstrate the generalizability of our method. Sec VII. concludes the whole paper.

\section{Related work}
\subsection{Salient Object Detection}
Salient Object Detection (SOD) has been studied as an important computer vision task for a long period of time. Traditional SOD methods \cite{Xia,itti1998model,jiang2013salient,yan2013hierarchical}  were mostly based on hand-crafted features. These low-level feature-based methods tend to work ineffectively when facing cluttered scenes. With the rise of deep learning, fully convolutional networks (FCNs)-based methods can not only utilize the low-level features of images, but also learn the ability to parse complex context of scenes from large scale datasets, greatly enhancing the robustness of SOD methods. Most of them are based on a backbone (\eg ResNet \cite{he2016deep}, VGG \cite{vgg}, \etal) pre-trained on ImageNet \cite{ImageNet} as feature encoders, varying the unique structures or modules under the paradigm of feature pyramid networks (FPNs \cite{lin2017feature}) to utilize multi-level features. A typical category of methods modifies the stream of high-level features based on the FPN architecture to improve performance. For example, Hou \etal \cite{hou2017deeply} introduce short connections to skip-layer structure in FPN architecture to better locate salient region and refine irregular prediction maps. Liu \etal \cite{liu2019simple} and Chen \etal \cite{chen2020global} design contextual modules to further capture global information of deep features and then use them to guide low-level features. Besides, several others add branches to introduce additional cues to help locate salient object and improve their boundary quality. Wei \etal \cite{wei2020label} decouple the ground-truth maps into interior and edge regions to separately supervise two branches and then aggregate them. Zhao \etal \cite{zhao2021complementary} design trilateral decoder including semantic path, spatial path and boundary path to solve the dilution of semantics, loss of spatial information and absence of boundary information. Furthermore, with the success of Transformer in vision tasks, Liu \etal \cite{liu2021visual} propose VST, which is the first unified model based on pure transformer for both RGB and RGB-D SOD task. Ma \etal \cite{BBNet} explore the combination of Transformer and CNN architecture, achieving state-of-the-art performance. Despite their impressive achievements on low-resolution SOD tasks, applying them directly to high-resolution inputs would be ineffective. On the one hand, the network is unable to capture global information due to the small receptive field relative to large resolution inputs, and on the other hand, it is hard to recover lost details if the inputs are down-sampled. Therefore, most of low-resolution methods cannot be directly applied to high-resolution scenes.

\subsection{High-resolution Salient Object Detection}
Nowadays, with the common access to high-resolution images, focusing on high-resolution SOD is already trending. In response to the challenges posed by high-resolution SOD researchers have made many contributions in terms of methods and datasets.

\textbf{High-resolution SOD datasets.}
DAVIS-S \cite{zeng2019towards} contains 92 images suitable for saliency detection from video object detection dataset DAVIS \cite{DAVIS}, which is precisely annotated and have a high resolution of $1920 \times 1080$. And all of the images are used for evaluation. Zeng \etal \cite{zeng2019towards} collect and annotate the first high-resolution saliency detection dataset, named HRSOD. It contains 1,610 and 400 images for training and testing, respectively. All images are collected from the website of Flickr. 40 participants are involved in pixel-level annotating. However, prior to our work this was the only dataset specific to high-resolution SOD training, which is insufficient compared to the low-resolution SOD dataset (\eg DUT-TR\cite{DUTS}, containing 10,553 images) for data-driven deep learning-based methods. Thus the goal of our UHRSD dataset is to support deep model learning to the more challenging details of high-resolution input and to provide comprehensive evaluation of methods performed on high-resolution conditions. More details and comparisons can be referred to \cref{dataset}.

\textbf{High-resolution SOD methods.}
Zeng \etal \cite{zeng2019towards} propose a paradigm for high-resolution salient object detection using GSN for extracting semantic information, and APS guided LRN for optimizing local details and finally GLFN for prediction fusion. Similarly, Tang \etal \cite{tang2021disentangled} disentangle the SOD task into two tasks. They first design LRSCN to capture sufficient semantics at low resolution and generate the tri-maps. By introducing the uncertainty loss, the designed HRRN can refine the tri-maps generated in first stage using low-resolution datasets. Both of them use multi-stage architecture, which lead to semantic incoherence between networks and slow inference. Besides, Zhang \etal \cite{DRFNet} propose DRFNet which consists of shared feature extractor and two effective refinement heads, adopting a global-aware feature pyramid and hybrid dilated blocks respectively. Recently, Deng \etal \cite{RMFormer} recurrently utilizes shared Transformers and multi-scale refinement architectures. In this way, high-resolution saliency maps can be generated with guidance of lower-resolution predictions. Although they are one-stage framework, treating large and small size of features equally leads to much slower inference. 

Unlike existing methods, we consider the advantages of CNNs and Transformers to design an asymmetric architecture. We use light-weight CNN to capture spatial features from large inputs and Transformer to capture contextual features from regular inputs, which not only can optimize the computational burden, but also the features extracted from heterogeneous encoders will form complementary effects. The novel grafting framework can fully leverage the advantages of heterogeneous encoders to better meet the demand of fast and accurate segmentation for real-life applications.

\begin{figure}[t]
    \centering
    \includegraphics[width=\linewidth]{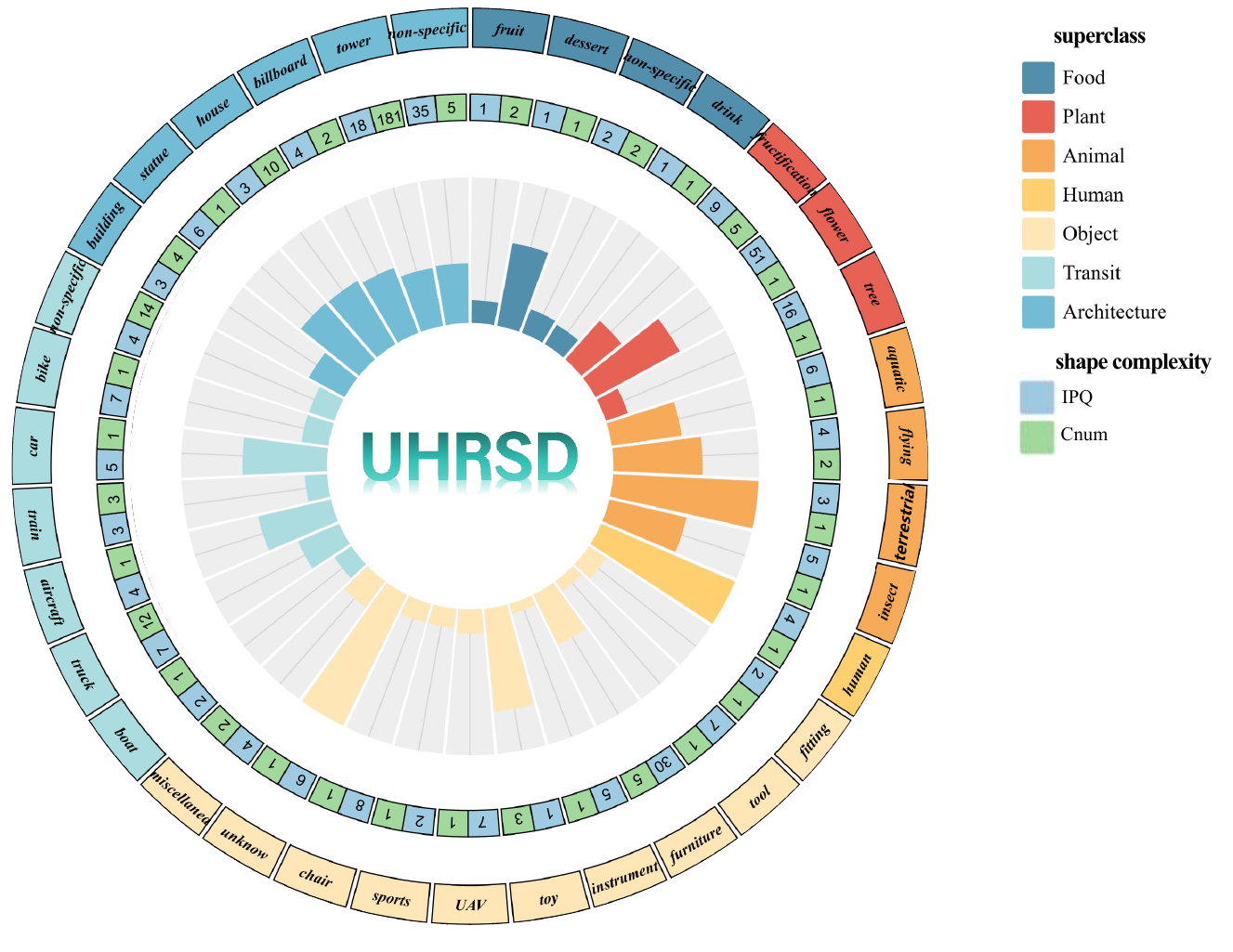}
    \caption{\textbf{Categories of our UHRSD dataset.} We illustrate 7 primary categories. For items within primary categories that are hard to classify, we categorize them as "non-specific". The columns' heights in this chart approximate the distribution of quantities across categories.}
    \label{fig:category}
\end{figure}

\section{Ultra High-Resolution Saliency Detection Dataset} \label{dataset}

\begin{table*}[t]
\caption{Comparison of statistics with existing SOD datasets. The table outlines and compares dataset attributes including type, number, image dimensions, and complexity levels. Te and Tr for Testing and Training respectively, optimal metrics shows in \textbf{Bold}.}
\label{table1}
\centering
\small
\fontsize{8pt}{11pt}\selectfont
\begin{tabular*}{1\textwidth}{@{\extracolsep{\fill}} p{2cm}|c|P{0.8cm}|P{2.1cm}|P{2.1cm}|P{1.8cm}|P{1.8cm}|P{1.8cm}} 
\toprule

\multicolumn{1}{c|}{\multirow{2}{*}{\begin{tabular}[c]{@{}c@{}}Dataset\end{tabular}}} & \multirow{2}{*}{Type} & Number & \multicolumn{2}{c|}{Dimension} & \multicolumn{3}{c}{Complexity} \\ \cline{3-8} 
 &  & $I_{num}$ & $H \pm \sigma_{H}$& $W \pm \sigma_{W}$ & $IPQ \pm \sigma_{IPQ}$ & $C_{num} \pm \sigma_{C}$ & $E_{num} \pm \sigma_{E}$ \\ \hline
SOD & \multicolumn{1}{l|}{Te} & \multicolumn{1}{l|}{300} & \multicolumn{1}{l|}{$366.87 \pm 72.35$} & \multicolumn{1}{l|}{$435.13 \pm 72.35$} & \multicolumn{1}{l|}{$5.98 \pm 5.22$} & \multicolumn{1}{l|}{$1.58 \pm 1.11$} & \multicolumn{1}{l}{$ \phantom{-} 0.92 \pm 1.67$} \\
PASCAL-S & \multicolumn{1}{l|}{Te} & \multicolumn{1}{l|}{850} & \multicolumn{1}{l|}{$387.63 \pm 64.65$} & \multicolumn{1}{l|}{$467.82 \pm 61.46$} & \multicolumn{1}{l|}{$4.04 \pm 3.00$} & \multicolumn{1}{l|}{$1.17 \pm 0.69$} & \multicolumn{1}{l}{$-2.79 \pm 11.75$} \\
ECSSD & \multicolumn{1}{l|}{Te} & \multicolumn{1}{l|}{1000} & \multicolumn{1}{l|}{$311.11 \pm 56.27$} & \multicolumn{1}{l|}{$375.45 \pm 47.70$} & \multicolumn{1}{l|}{$4.05 \pm 3.21$} & \multicolumn{1}{l|}{$1.12 \pm 0.46$} & \multicolumn{1}{l}{$\phantom{-}0.55 \pm 1.36$} \\
HKU-IS & \multicolumn{1}{l|}{Tr+Te} & \multicolumn{1}{l|}{4447} & \multicolumn{1}{l|}{$292.42 \pm 51.13$} & \multicolumn{1}{l|}{$386.64 \pm 37.42$} & \multicolumn{1}{l|}{$5.45 \pm 4.88$} & \multicolumn{1}{l|}{$1.68 \pm 1.17$} & \multicolumn{1}{l}{$\phantom{-}1.14 \pm 1.91$} \\
DUT-OMRON & \multicolumn{1}{l|}{Te} & \multicolumn{1}{l|}{5168} & \multicolumn{1}{l|}{$320.93 \pm 54.35$} & \multicolumn{1}{l|}{$376.78 \pm 46.02$} & \multicolumn{1}{l|}{$4.44 \pm 5.12$} & \multicolumn{1}{l|}{$1.30 \pm 1.01$} & \multicolumn{1}{l}{$\phantom{-}0.34 \pm 3.27$} \\
DUTS & \multicolumn{1}{l|}{Tr+Te} & \multicolumn{1}{l|}{\textbf{15572}} & \multicolumn{1}{l|}{$322.10 \pm 53.69$} & \multicolumn{1}{l|}{$375.48 \pm 47.03$} & \multicolumn{1}{l|}{$3.76 \pm 3.89$} & \multicolumn{1}{l|}{$1.49 \pm 12.19$} & \multicolumn{1}{l}{$\phantom{-}0.12\pm 4.03$} \\
SOC & \multicolumn{1}{l|}{Te} & \multicolumn{1}{l|}{3000} & \multicolumn{1}{l|}{$480.00 \pm 0.00$} & \multicolumn{1}{l|}{$640.00 \pm 0.00$} & \multicolumn{1}{l|}{$4.61 \pm 3.79$} & \multicolumn{1}{l|}{$1.77 \pm 1.72$} & \multicolumn{1}{l}{$-10.09 \pm 29.81$} \\
XPIE & \multicolumn{1}{l|}{Te} & \multicolumn{1}{l|}{10000} & \multicolumn{1}{l|}{$399.44 \pm 69.01$} & \multicolumn{1}{l|}{$464.72 \pm 63.46$} & \multicolumn{1}{l|}{$4.19 \pm 3.53$} & \multicolumn{1}{l|}{$1.17 \pm 0.50$} & \multicolumn{1}{l}{$\phantom{-}0.56 \pm 1.63$} \\
MSRA10K & \multicolumn{1}{l|}{Te} & \multicolumn{1}{l|}{10000} & \multicolumn{1}{l|}{$324.51 \pm 56.26$} & \multicolumn{1}{l|}{$370.27 \pm 50.25$} & \multicolumn{1}{l|}{$3.12 \pm 2.94$} & \multicolumn{1}{l|}{$1.07 \pm 0.53$} & \multicolumn{1}{l}{$-1.89 \pm 17.60$} \\
DAVIS-S & \multicolumn{1}{l|}{Te} & \multicolumn{1}{l|}{92} & \multicolumn{1}{l|}{$1299.13 \pm 440.77$} & \multicolumn{1}{l|}{$2309.57 \pm 783.59$} & \multicolumn{1}{l|}{$\mathbf{7.38 \pm 5.17}$} & \multicolumn{1}{l|}{$2.84 \pm 6.05$} & \multicolumn{1}{l}{$-9.92 \pm 20.37$} \\
HRSOD & \multicolumn{1}{l|}{Tr+Te} & \multicolumn{1}{l|}{2010} & \multicolumn{1}{l|}{$2713.12 \pm 1041.70$} & \multicolumn{1}{l|}{$3411.81 \pm 1407.56$} & \multicolumn{1}{l|}{$5.21 \pm 4.79$} & \multicolumn{1}{l|}{$1.74 \pm 2.33$} & \multicolumn{1}{l}{$-2.70 \pm 14.74$} \\
UHRSD & \multicolumn{1}{l|}{Tr+Te} & \multicolumn{1}{l|}{5920} & \multicolumn{1}{l|}{{$\mathbf{3718.60 \pm 850.28}$}} & \multicolumn{1}{l|}{$\mathbf{4943.79 \pm 1026.38}$} & \multicolumn{1}{l|}{$6.77 \pm 8.92$} & \multicolumn{1}{l|}{$\mathbf{3.19 \pm 9.16}$} & \multicolumn{1}{l}{$\mathbf{-15.05 \pm 121.21}$} \\ \bottomrule
\end{tabular*}
\end{table*}
The existing common SOD datasets usually are in low-resolution (below 500 × 500 ), which present following drawbacks for training high-resolution networks and evaluating high-quality segmentation results. Firstly the low resolution of the images results in insufficient detail information. Secondly, the edge quality of annotations is poor. Lastly, the finer level of annotations is frequently inadequate, especially for hard-case annotations which are handled perfunctorily as shown in \cref{fig:maskquality}. Before this, the only available high-resolution dataset known is HRSOD \cite{zeng2019towards}. However, the number of high-resolution images in HRSOD is limited, training exclusively on it tend to cause over-fitting its specific data distribution, which significantly impacted the model’s generalization ability.  With this in mind, our goal is to contribute a larger and more challenging high-resolution dataset UHRSD for tackling the lack of \textbf{training and evaluating dataset in high-resolution SOD domain}. We will provide more details in Dataset Collection and Annotation, and Dataset Analysis.

\begin{figure}
    \centering
    \includegraphics[width=\linewidth]{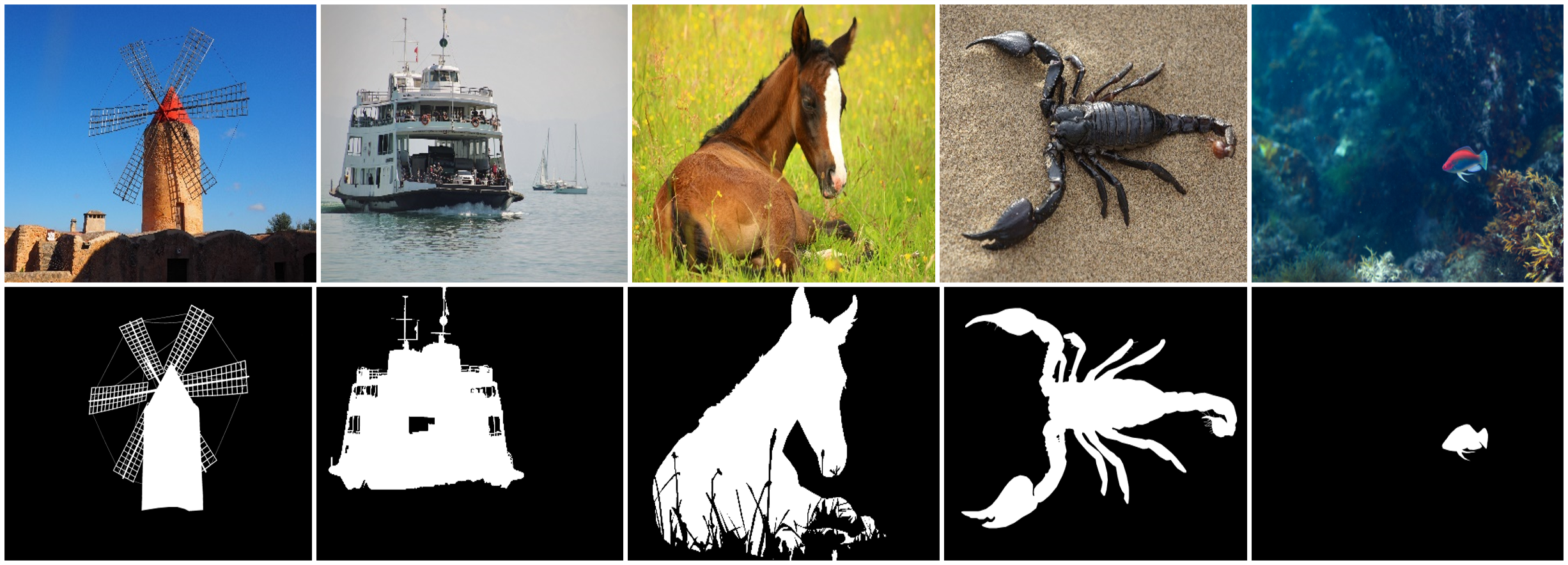}
    \caption{\textbf{Examples and corresponding annotations in UHRSD.} UHRSD contains rich salinent objects in terms of classes and attributes. }
    \label{fig:uhrsd_sample}
\end{figure}

\subsection{Dataset Collection and Annotation}
Considering that a single dataset is unlikely to perfectly encompass all aspects, having a clear objective and defined criteria during the collection process is crucial for dataset construction. In alignment with our goal, our construction criteria for UHRSD are threefold: 

(1) adhering to the SOD dataset standards, which encompass basic characteristics such as contrast, center bias and \etc; (2) ensuring diversity in categories to cover common salient objects, thereby enhancing the model's generalization capability; (3) taking into account high-resolution image attributes, which guarantees higher resolution and more complex shape challenges compared to existing datasets.
Adhering to our criteria, we have gathered 5,920 images, all surpassing 4K resolution and featuring a variety of salient objects. Most images were sourced from websites like Flickr and Pixabay and applied for academic use.

During the collection process, to ensure the diversity of salient objects, we initially analyzed the categories present in the widely used DUTS and then we devised keywords including \textit{food, animals, architecture, humans, \etc} for the preliminary image gathering. 
Subsequently, multiple participants were involved in meticulously selecting images with noticeable salient objects, with an emphasis on selecting those with complex shapes. The categories and proportions of the final image set are illustrated in \cref{fig:category}.

\begin{figure*}[t]
    \centering
    \includegraphics[width=\textwidth]{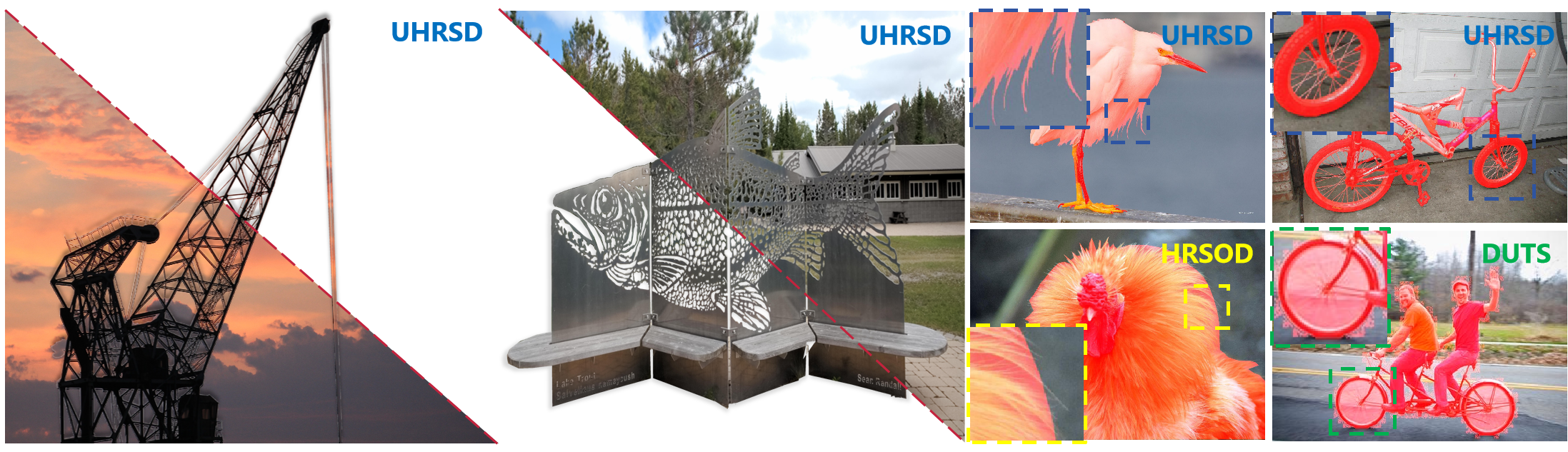}
    \caption{\textbf{Comparison of annotation quality among UHRSD and other SOD datasets.} From left to right: 2 sample images from UHRSD; comparison of annotation quality between UHRSD and HRSOD; comparison of annotation quality between UHRSD and DUTS. {Best viewed by zoom-in.}}
    \label{fig:maskquality}
\end{figure*}

During the data labeling stage, we employed labeling standards of precision far exceeding those of previous datasets. Participants initially identified salient objects through eye-tracking experiments, followed by pixel-level annotation. Examples from UHRSD are shown in \cref{fig:uhrsd_sample}. Notably, for transparent or hollow areas, which were not distinguished in prior datasets, we have accurately labeled them as shown in \cref{fig:maskquality}.

\begin{figure}[]
    \centering
    \includegraphics[width=\linewidth]{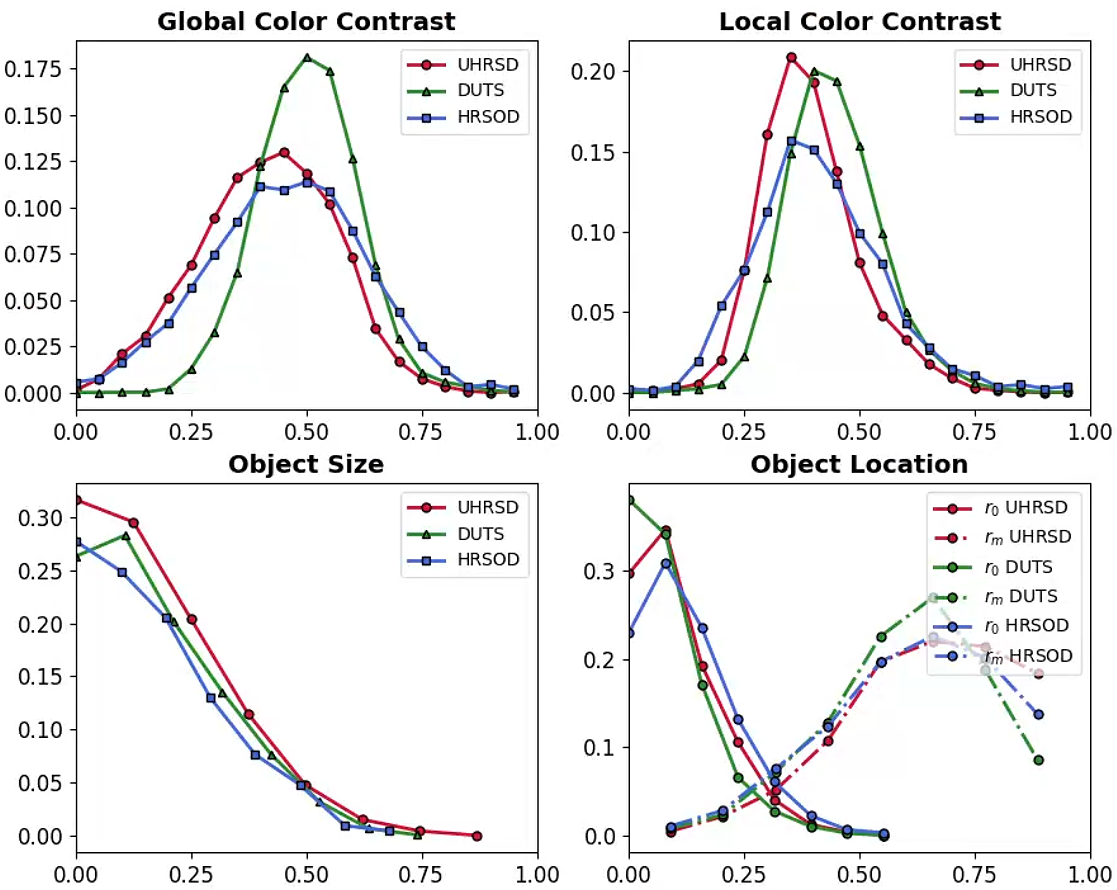}
    \caption{\textbf{Comparison of basic attributes among UHRSD, HRSOD and DUTS datasets.} \textit{Top-left/right} show the global/local color contrast distribution, aligning with the standards of SOD datasets. \textit{Bottom-left} shows comparison of object size distribution in UHRSD. And the \textit{bottom-right} displays the distribution of object locations, reflecting the center-bias of these datasets. } 
    \label{fig:attribute}
\end{figure}

\subsection{Dataset Analysis}
Our comprehensive comparison of UHRSD encompasses an in-depth analysis of its multiple aspects, as outlined in \cref{table1}. And details are as follows:
\begin{itemize}
    \item  \textbf{Resolution} The fourth and fifth columns in \cref{table1} display the  average resolution and variance of commonly used SOD datasets, indicating that most SOD datasets have a small resolution, making it challenging to include sufficient details. The UHRSD dataset, with its average dimensions meeting the 4K standard, significantly surpasses previous datasets like HRSOD and low-resolution SOD datasets. Its ultra-high resolution not only meets current demands but also expand opportunities for future high-resolution segmentation research.

    \item \textbf{Color contrast} In accordance with \cite{contrast}, we present the global/local contrast to affirm that our collected images are salient in terms of low-level vision. Top-left/right in \cref{fig:attribute} reveals that the contrast intensity of the UHRSD dataset cluster around the median, aligning with the normative range for SOD datasets. The contrast intensity of UHRSD is slightly lower than that of DUTS, suggesting that UHRSD poses a greater challenge to some extent.
    
    \item \textbf{Object size} We utilize the foreground-to-backgroud area ratio to demonstrate the object size. Bottom-left in \cref{fig:attribute} shows that UHRSD offers a more varied distribution of object sizes, including a wide range of salient object sizes.

    \item \textbf{Center bias} As shown in \cref{fig:attribute} \textit{bottom-right}, we adopt distances between object center, object margin and the image center to explain the center bias of the datasets. Upon comparison, it is evident that the center bias distribution of UHRSD is similar to that of commonly used SOD datasets, aligning with the characteristic center bias of SOD datasets.
    
    \item \textbf{Shape complexity} Inspired by \cite{DIS}, in the last 3 columns of \cref{table1}, we measure the complexity of salient objects using three metrics: \textit{isoperimetric inequality quotient}($IPQ$), \textit{number of object contours}($C_{num}$)\cite{DIS} and \textit{Euler number}($E_{num}$)\cite{Eulernumber}. The $IPQ$ is calculated based on the isoperimetric inequality $IPQ=\frac{C^2}{4\pi A}$, where $C$ and $A$ denotes total perimeter and area respectively, measuring the overall compactness of the shape. $C_{num}$ reflects the complexity of the contour, with higher values indicating more intricate edges. $E_{num}$ represents the difference between the number of objects and holes, with smaller values indicating a greater number of holes than objects. From the Complexity columns in the table, it's evident that the complexity metrics of the UHRSD datasets considerably surpass those of existing datasets. Compared to the HRSOD dataset, the metrics of $IPQ$, $C_{num}$, $E_{num}$ in UHRSD are higher by $29.9\%$, $83.3\%$ and $457.4\%$, respectively. This indicates that UHRSD better utilizes the advantages of high-resolution images and encompasses more complex details, thereby making it more challenging and valuable for research in HR-SOD field.
    
    \item \textbf{Mask quality} \cref{fig:maskquality} shows some samples from our UHRSD dataset and other HR/LR-SOD datasets. The first two images in \cref{fig:maskquality} demonstrate the complexity of images in UHRSD, all of which are carefully annotated with great manual effort. The four sub-images on the right illustrate UHRSD's superior annotation quality by comparing it with similar mask types from existing HR-/LR-SOD datasets. For instance, in the comparison with HRSOD, the mask in HRSOD represents complex hair edges with approximate contours, whereas ours precisely follows the actual boundaries. In addition, our high quality is not only reflected in the precision. When contrasted with LR-SOD dataset DUTS, it is apparent that our annotations are of a higher standard, accurately dealing with challenges like hollow and obscured areas, which are always treated simply.

\begin{figure*}[t]
    \centering
    \includegraphics[width=\textwidth]{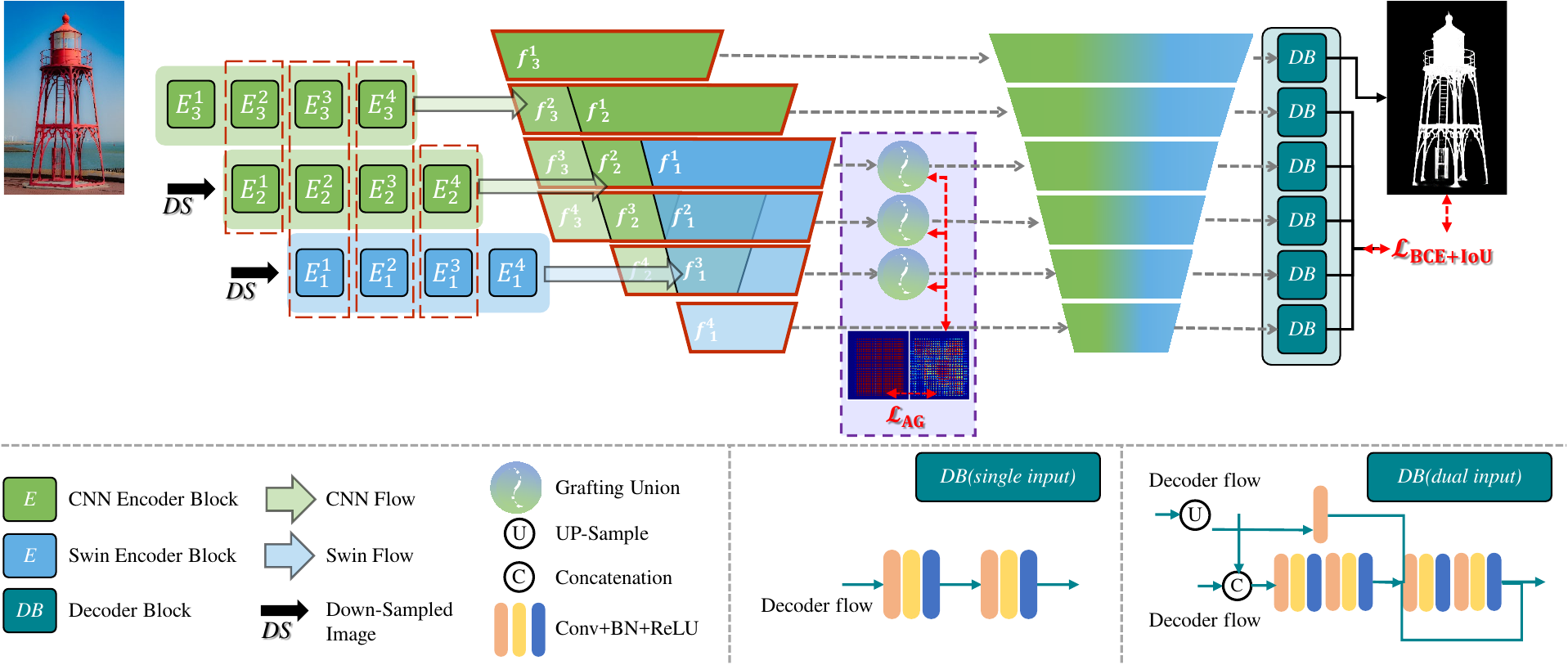}
    \caption{\textbf{Pipeline of our PGNeXt framework.} Three parallel encoders are set up, the two indicated in green are ResNet-18-based and the one indicated in blue at the bottom is Swin Transformer-based. The three feature groups extracted by corresponding encoders  are connected in a staggered manner and grafted through the grafting union to construct a higher feature pyramid. The grafted features are decoded from the bottom up by the decoder. The decoder blocks that compose the decoder are all dual-input except the bottom one which is single-input.}
    \label{fig:3}
\end{figure*}

    \label{sec:dataset}
    \item \textbf{Dataset splitting}
    For ease of use, we randomly divided 5,920 images into a training set of 4,932 images (\ie UHRSD-TR) and a test set of 988 images (\ie UHRSD-TE), maintaining a roughly 5:1 ratio. This division ensures consistency in the categories between the training a test set. Given that SOD is a category-agnostic segmentation task and HR-SOD places a higher emphasis on the challenge of segmenting high-resolution details, we opted to split the test set based on the shape complexity. Therefore, inspired by \cite{DIS} we sorted the 988 test images into four equally sized subsets (\ie UHRSD-TE 1-4) in ascending order according to their $IPQ \times C_{num} $ scores. Testing on these four subsets, each representing a different level of shape complexity, allows for a more comprehensive assessment of the SOD model's ability to segment complex objects. The detailed quantitative evaluation for the subsets can be found in \cref{sec:subsets}. 
\end{itemize}

\section{PGNeXt}
\subsection{Network Overview}
Feature pyramid networks (FPNs) have been widely used as a classic architecture and have proven to be effective for SOD tasks \cite{zhao2021complementary, hou2017deeply, wei2020f3net}. This U-shaped architecture first extracts a series of features from the bottom up and then fuses the multi-scale features in the top-down path. It is widely acknowledged that the low-level features preserve rich spatial features helpful for recovering the local details in saliency maps, and the high-level features contain contextual semantics conducive to the accurate identification of salient object. However, the fact that this architecture cannot be directly applied on high-resolution SOD tasks can be attributed to the contradiction of two factors. On the one hand, during bottom-up feature encoding, the relative receptive field to the high-resolution feature maps is limited to extracting accurate high-level semantics. On the other hand, if we down-sample the input images to ensure the relative receptive field is enough, we cannot retain rich spatial features to recover the details during the top-down decoding process. Therefore, the main motivation of our method is to resolve the paradox in traditional FPNs architecture for HR-SOD task.

The pipeline of our method shows in \cref{fig:3}. The pipeline can be regarded as three phases, pyramid feature extraction stage, pyramid feature grafting stage and pyramid feature decoding stage. Firstly, in the pyramid feature extraction stage, for the contradiction we adopt two strategies, i.) set up three parallel encoders to extract features from inputs with multiple resolutions, so that diverse features contain both low-level details and high-level semantics can be retained. ii.) to further obtain global contexts and rich details, we simultaneously use Transformer-based and CNN-based encoders to extract diverse features for following grafting stage. In grafting stage, for the extracted group of features we adopt hierarchical staggered grafting framework and introduce the window-based Cross-Model-Grafting module (wCMGM) to progressively make semantic information transfer between parallel branches, so as to build a higher feature pyramid. Besides, the wCMGM generates a matrix named cross-attention matrix (CAM) to be supervised by Attention Guided Loss (AGL) for enhancing the cross attention in wCMGM. Finally, the constructed higher feature pyramid is decoded from coarse to fine to recover high-resolution details and we can get the final high-resolution saliency map. The final prediction and side outputs from each decoder block are supervised by the ground-truth maps. In what follows, we elaborate the details of the hierarchical staggered grafting framework, window-based Cross-Model Grafting modules and objective functions. 

\subsection{Hierarchical Staggered Grafting Framework}
\textbf{Feature Extractor.} For input image $I \in \mathbb{R}^{3 \times H \times W}$, we can obtain a set of images of a specific resolution $\{I_i \in \mathbb{R}^{3 \times H/2^{3-i} \times W/2^{3-i}} \}_{i=1}^3$by bilinear sampling. Correspondingly, we set up three parallel branches to extract three sets of features with different resolutions, which can be denoted as $\{E_i\}_{i=1}^3$. Based on the motivation mentioned above, we want to extract the global information while preserving it as much detail as possible during the feature extraction. Further, inspired by the existing research, transformer has strong global modeling capability due to its attention mechanism, while CNN can better preserve spatial details. Therefore, in order to balance the effectiveness and efficiency, for the smallest resolution branch $E_1$ we adopt transformer-based encoder to obtain the global context of the image, whereas for the larger resolution branch $E_2$ and $E_3$ we adopt light-weight CNN-based encoder to reduce
the computational burden while extracting large-size spatial features. More specifically, we choose the representative Swin Transformer and ResNet-18 as transformer-based encoder and CNN-based encoder respectively. For ResNet-18, the feature map extracted by top $7 \times 7$ layer offers limited performance gains but consume huge computational effort, especially for high-resolution input. Thus we only adopt the features extracted in last four stages in ResNet-18, which can be denoted as $\{f_i^j \}_{i=2,3}^{j=1,2,3,4}$, where $i$ denotes the $i$-th encoder and $j$ denotes the $j$-th encoder stage. For Swin Transformer, since the resolution is same in last two stages, we simply fuse the output features of them as the last output feature, while adopting the patch embedding feature, which generates 4 features denoted as $\{f_i^j\}_{i=1}^{j=1,2,3,4}$. 

\textbf{Hierarchical Staggered Connection.} 
Although three groups of features from three parallel branches were obtained, covering the low-level details and high-level semantics of the high-resolution images, there are gaps between branches and layers, resulting in the failure to directly aggregate for prediction. To this end, we designed the hierarchical staggered connection to keep high resolution of feature maps and to allow global semantics to transfer gradually from low-resolution branch to higher-resolution branch, so that the high-resolution features which lack semantic due to insufficient receptive fields, are organized by the global contextual information. In particular, for feature $f_i^j$, its dimensions satisfy 
\begin{equation} \label{eq1}
    f_i^j \in \mathbb{R}^{C_i^j \times H/2^{4-i+j} \times W/2^{4-i+j}}.
\end{equation}

It can be inferred from \cref{eq1} that for feature $f_i^j$ and $f_{i+1}^{j-1}$, they are same in spatial resolution, which implies that they may be similar for spatial representations. Besides, these two features pass through network layers of approximate depth, which leads to a minor semantic discrepancy. Thus the way we use the spatial dimension as clues constitutes a kind of staggered connection, and further through the wCMGM the global semantic information is grafted from low-resolution branch to higher-resolution branch. Up to this point, we construct a higher feature pyramid through three lower pyramid using hierarchical staggered connection structure as shown in \cref{fig:3}. In other word, the network achieves deeper sampling depth at low computational cost to adapt to the challenge caused by high-resolution input.

\begin{figure}[t]
    \centering
    \includegraphics[width=\linewidth]{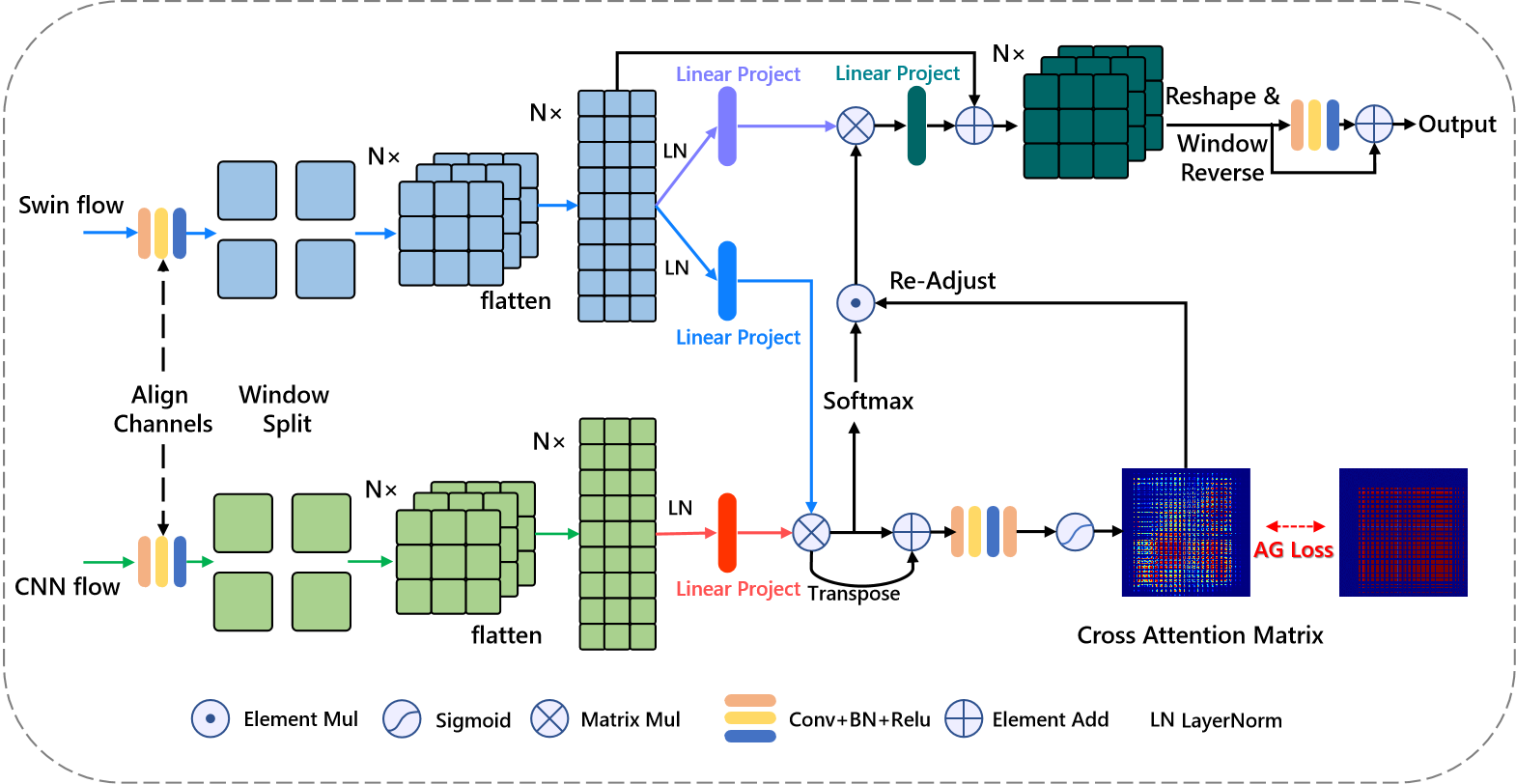}
    \caption{Architecture of window-based Cross-Model Grafting Module.}
    \label{fig:wcmgm}
\end{figure}

\subsection{window-based Cross-Model Grafting Module}
We propose wCMGM to graft the features extracted by different types of encoders. For feature extracted by Swin backbone, due to the transformer's ability to capture information over long distance, it has global semantic information that is important for saliency detection. In contrast, CNNs perform well at extracting local information thus features $\{f_i^j\}_{i=2,3}^{j=1,2,3,4}$ have relatively rich details. However, due to the contradiction between feature size and receptive field, there will be many noises in the background area of the saliency maps. For a salient prediction of a certain region, the predictions generated from different features can be roughly summarized as three cases : (a) Both right, (b) Some of them right and (c) Both wrong.

Existing fusion techniques, such as addition and multiplication, utilize element-wise convolutional operations that only can focus on limited local information, fail to address common errors in fusion methods. Therefore, they are effective only in the first two cases. 
Compared with the existing feature fusion, wCMGM recalculates the point-wise relationship between CNN feature and Transformer feature, transferring the global semantic information from Transformer branch to CNN branch so as to remedy the common errors. Since we aim to transfer global information to local branches rather than re-modeling global relations on the feature map, we can compute within non-overlap window, and we found that this not only reduces the computation significantly but also improves the performance further.

Specifically, we first use two successive convolutional layers followed by batch normalization layer and ReLU layer for compressing the channel number of features from different branches. After that, we use element-wise addition merge inputs from ResNet-18 branch. As shown in \cref{fig:wcmgm}, we denote the aggregated feature from ResNet and feature from Swin as $f_R \in \mathbb{R}^{H\times W \times C}$ and $f_S \in \mathbb{R}^{H\times W \times C}$ respectively. Then we partition the features in a non-overlapping manner into $N$ windows of size $S \times S$, where $N$ equals to $\frac{HW}{S^2}$. For $f_R^i \in \mathbb{R}^{S\times S\times C}$, we flatten it to $f_R^{i'} \in \mathbb{R}^{1\times C \times S^2}$ and do the same to $f_S^i$ to get $f_S^{i'}$. Inspired by the multi-head self-attention mechanism, we apply layer normalization and linear projection on them respectively to get ${f_R^i}^q$, ${f_R^i}^v$ and ${f_S^i}^k$. We can get the attention matrix $Y$ by matrix multiplication, which can be denoted as
\begin{equation}
    Y^i = ({f_R^i}^q \times {{f_S^i}^k}^T).
\end{equation}
Next, we generate the cross attention matrix CAM based on $Y$, which can be expressed as
\begin{equation}
    \mathrm{CAM}^i = \sigma(\mathrm{BN}(\mathrm{Conv}(Y^i+{Y^i}^T))),
\end{equation}
where $\mathrm{Conv}$ and $\mathrm{BN}$ denote convolutional layer and batch normalization respectively, and $\sigma(\cdot)$ is the \textit{sigmoid} function. 
The generated $\mathrm{CAM}$ is supervised with proposed Attention Guided Loss (AGL) in order to readjusting the cross attention in wCMGM. Inspired by \cite{yu2022democracy}, the readjusting coefficient ${Y^i}^\text{re}$ and enhanced attention map ${Y^i}^\text{final}$ can be obtained from 
\begin{equation}
    {Y^i}^\text{re} = (\mathrm{CAM^i}+1)^\alpha,
    \label{eq:alpha}
\end{equation}
\begin{equation}
    {Y^i}^\text{final} = \mathrm{softmax}(Y^i) \odot {Y^i}^\text{re},
\end{equation}
where $\alpha$ is hyper-parameter used for adjusting the amplification rate and $\odot$ is t he element-wise multiplication. The salient features should have a higher similarity, thus the dot production in $\mathrm{CAM^i}$ should have a larger activation value. Therefore, ${Y^i}^{re}$ will amplify the the positive value at these points in final attention map to further strengthen the ability of selecting discriminating features in cross attention. We obtain $Z^i$ by matrix multiplication as follow:
\begin{equation}
    Z^i = {Y^i}^\text{final} \times {f_R^i}^v,
\end{equation}
then we input $Z^i$ to the linear projection layer and reshape it to the size of $\mathbb{R}^{S\times S\times C}$. After processing all windows we merge the obtained $\{Z^i\}_{i=1}^N$. Two shortcut connections were performed in the process as shown in \cref{fig:wcmgm} and we can get the grafted feature.

\begin{figure}[t]
    \centering
    \includegraphics[width=0.85\linewidth]{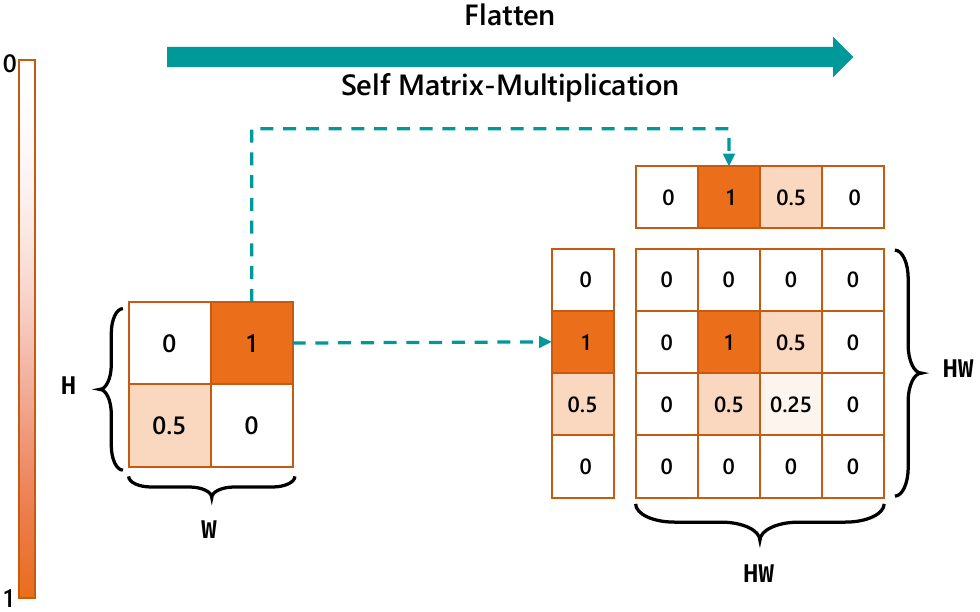}
    \caption{The construction of attention matrix. The operation is used to create target and weights for proposed AGL.}
    \label{fig:agl}
\end{figure}

\subsection{Attention Guided Loss}
In order for wCMGM to better serve the purpose of transferring information from the Transformer branch to the CNN branch, we design the Attention Guided Loss (AGL) to explicitly supervise the CAM to enhance the attention in wCMGM. We argue that the CAM should be similar to the attention matrix generated from ground truth, because the salient features should have a higher similarity resulting in higher dot product value in CAM. Through supervising the CAM and using it as an enhancement coefficient, the value of specific locations in the attention matrix so that the features of salient regions are emphasized. As shown in \cref{fig:agl} given a salient map $M$ with size $H \times W$, we first flatten it to $M'$ to obtain corresponding attention matrix $M^a$. The process can be denoted as $M^a=\mathcal{F} (M)$ and the value of $M^a_{xy}$ can be expressed as 

\begin{equation}
    M_{xy}^a = {M^i}_x^T \times M'_y 
\end{equation}

Then we use the transformation $M^a=\mathcal{F} (\cdot)$ to construct $G^a$, where $G$ is the ground truth map. Before applying the $\mathcal{F}(\cdot)$, we split it into $N$ non-overlapping windows just like the manner in wCMGM. We propose the AGL based on binary cross entropy (BCE) to supervise the Cross Attention Matrix generated in wCMGM shown in \cref{fig:wcmgm}. The BCE\cite{de2005tutorial} can be written as
\begin{equation}
    \ell_\text{BCE}(G_{xy}, P_{xy}) = \begin{cases}
    \log(P_{xy})  &{G_{xy}=1}\\
    \log(1-P_{xy}) &{G_{xy}=0}
    \end{cases},
\end{equation}
where $G_{xy}$ is the ground truth label of the pixel $(x,y)$, and $P_{xy}$ is the predicted probability in predicted map and both of them are in range $[0,1]$. Then our $\ell_\text{AG}$ can be expressed as
\begin{equation}
    \ell_\text{AG}(G,\mathrm{CAM}) = \frac{\sum\limits^N_{k=1}\sum\limits_{i,j}^{S,S}{\ell_\text{BCE}(G_{ij}^k,\mathrm{CAM}^k_{ij})}}{N},
\end{equation}
where $N$ equals to the number of windows and $S$ is the window size.
What's more, we also apply the widely-used intersection-over-union (IoU) loss \cite{mattyus2017deeproadmapper} to pay more attention to the global structure of the images as suggested by \cite{qin2019basnet}. The IoU loss $\mathcal{L}_\text{IoU}$ can be expressed as 
\begin{equation}
    \ell_\text{IoU}(G,P)=1-\frac{\sum\limits^{H,W}_{i,j}(G_{ij} \times P_{ij})}{\sum\limits^{H,W}_{i,j}(G_{ij}+P_{ij}-G_{ij} \times P_{ij})}.
\end{equation}
In the end, our total loss can be expressed as follow:
\begin{equation}
    \mathcal{L}_\text{total} = \sum\limits_{i=1}^6 \mathcal{L}_\text{BCE+IoU}^i+\sum \limits_{j=1}^3\mathcal{L}_\text{AG}^j,
\end{equation}
where $\mathcal{L}_\text{BCE+IoU}^i$ hybrid loss applied on the final prediction and other 5 side-outputs from decoder, and $\mathcal{L}_\text{AG}^j$ is used for supervising CAMs from 3 wCMGMs.

\section{HR-SOD benchmark}
\subsection{Experimental Settings}
\paragraph{Implementation Details} We use PyTorch\cite{Pytorch} to implement our model and one NVIDIA RTX 3090 GPU is used for both training and testing. For simplification, we utilize the ResNet-18 architecture for our CNN branches. It should be noted that we did not use pre-trained parameters for ResNet-18 encoders, allowing for the flexibility to replace the CNN branch with more efficient or more effective CNN encoders for better performance or speed. Swin-B \cite{Swin} is adopted as the transformer encoder to stay in line with the advanced HR-SOD methods\cite{RMFormer,InSPyReNet}. The whole network is trained end-to-end by using stochastic gradient descent (SGD). We set the maximum learning rate to 0.018, with the learning rate for Swin-B encoder frozen at 0.05 of the maximum learning rate. The learning rate first increases then decays during the training process, what's more momentum and weight decay are set to 0.9 and 0.0005 respectively. Batchsize is set to 8 and maximum epoch is set to 32.
\paragraph{Datasets} We use DUTS-TR \cite{DUTS}, HRSOD-TR \cite{zeng2019towards} and UHRSD-TR datasets for training process. For fair comparison with existing methods \cite{RMFormer} and alignment with our prior work\cite{PGNet}, we adopt two training setups: DUTS+HRSOD (\textbf{DH}) and HRSOD+UHRSD (\textbf{UH}). To demonstrate the performance on high-resolution images, we evaluate the models on DAVIS-S\cite{zeng2019towards}, HRSOD-TE\cite{zeng2019towards} and our UHRSD-TE datasets. 
\paragraph{Evaluation Metrics}
We use following 5 metrics widely used in SOD tasks to evaluate the performance of all methods: mean absolute error ($\mathcal{M}$), \textit{S}-measure ($S_\alpha$)\cite{fan2017structure}, \textit{E}-measure ($E_m$)\cite{fan2018enhanced}, weighted \textit{F}-measure ($F^w_\beta$) \cite{wfm} and Max \textit{F}-measure ($F^{M}_\beta$).

\begin{itemize}
    \item \textbf{$\mathcal{M}$}: The mean absolute error $\mathcal{M}$ refers to the pixel-level average difference between prediction map \textit{P} and ground truth \textit{G}, which is defined as
    \begin{equation}
        \label{eq:mae}
        \mathcal{M}=\frac{1}{H \times W}\sum\limits^H_{i=1}\sum\limits^W_{j=1}|P_{ij}-G_{ij}|.
    \end{equation}

    \item \textbf{S-measure} The S-measure evaluates region-aware and object-aware structural similarity between prediction maps and ground-truth by
    \begin{equation}
        \label{eq:sm}
        S_\alpha=\alpha \cdot S_o + (1-\alpha) \cdot S_r,
    \end{equation}
    where $\alpha=0.5$. $S_o$ and $S_r$ refer to region-aware and object-aware similarity measure respectively.
    \item \textbf{E-measure} The enhanced-alignment measure aligns with human visual perception by concurrently capturing pixel-level and image-level matching.
    \item \textbf{Max {F}-measure} The Max {F}-measure can be calculated by
    \begin{equation}
        \label{eq:fm}
        F_\beta = \frac{(1+\beta^2)\cdot \mathrm{precision }\cdot \mathrm{recall}}{\beta^2 \cdot \mathrm{precision } + \mathrm{recall}},
    \end{equation}
    where $\beta^2$ is set to 0.3 as suggested in \cite{borji2015salient}. The Max {F}-measure refers to the maximum $F_\beta$ value from all precision-recall pairs generated after applying thresholds ranging from 0 to 255.
    \item \textbf{weighted F-measure} The weighted F-measure follow the \cref{eq:fm}, replacing $\mathrm{precision}$ and $\mathrm{recall}$ with $\mathrm{precision}^w$ and $\mathrm{recall}^w$, which intuitively generalize the primitive F-measure.
\end{itemize}

In addition, we also report the metric of mean boundary accuracy (mBA) \cite{mBA} to further evaluate the boundary quality which is important in HR-SOD.

\begin{table*}[t]
\caption{Quantitative comparisons with state-of-the-art SOD models on three high-resolution benchmark datasets in terms of  max F-measure, weighted F-measure, MAE , E-measure, S-measure and mBA. LR methods: trained on DUTS-TR, -DH: trained on DUTS-TR and HRSOD-TR, -UH: trained on UHRSD-TR and HRSOD-TR . The best two results are highlighted in \textbf{bold} and \underline{underlined}.}

\label{table:performance}
\LARGE
\renewcommand\arraystretch{1.3}
\centering
\setlength\tabcolsep{6pt}
\resizebox{0.95\textwidth}{!}{%

\begin{tabular}{lccccccccccccccccccccccc}
\toprule[2pt]
\multicolumn{1}{l|}{} & \multicolumn{6}{c|}{\textbf{HRSOD-TE}}        & \multicolumn{6}{c|}{\textbf{DAVIS-S}} & \multicolumn{6}{c}{\textbf{UHRSD-TE}}                              \\ \cline{2-19} 
\multicolumn{1}{l|}{\multirow{-2}{*}{\textbf{Method}}} 
& $F_\beta^{M}$ & $F_\beta^{w}$ & $\mathcal{M}$ & $E_\xi$ & $S_\alpha$ & \multicolumn{1}{c|}{$\mathrm{mBA}$}
& $F_\beta^{M}$ & $F_\beta^{w}$& $\mathcal{M}$ & $E_\xi$& $S_\alpha$ 
& \multicolumn{1}{c|}{$\mathrm{mBA}$}  
& $F_\beta^{M}$ & $F_\beta^{w}$& $\mathcal{M}$ & $E_\xi$ & $S_\alpha$
& \multicolumn{1}{c}{$\mathrm{mBA}$}        
\\ \toprule[2pt]
\hline
\multicolumn{19}{c}{\textbf{LR-SOD methods}}                               \\ \midrule \hline
\multicolumn{1}{l|}{\textbf{DASNet}\cite{zhao2020depth}} 
& .894& .847 & .032 & .925 & .898 & \multicolumn{1}{c|}{.657}
& .902& .861 & .020 & .949 & .911 & \multicolumn{1}{c|}{.665}
& .916& .868 & .044 & .894 & .890 & \multicolumn{1}{c}{.676}
\\
\multicolumn{1}{l|}{\textbf{F3Net}\cite{wei2020f3net}} 
& .901 &.835 &.035 &.913 &.897& \multicolumn{1}{c|}{.661} 
& .915 &.848 &.020 &.940 &.914& \multicolumn{1}{c|}{.667} 
& .911 &.856 &.046 &.889 &.892 & \multicolumn{1}{c}{.684}
\\
\multicolumn{1}{l|}{\textbf{GCPA}\cite{chen2020global}} 
& .889&.816 & .036 & .898 & .898 & \multicolumn{1}{c|}{.656}   
& .922&.846 & .020 & .934 & .929& \multicolumn{1}{c|}{.663} 
& .912&.851  &.047 & .886 & .896& \multicolumn{1}{c}{.680}\\
\multicolumn{1}{l|}{\textbf{ITSD}\cite{zhou2020interactive}} 
& .896&.828 & .036 & .912& .898& \multicolumn{1}{c|}{.676} 
& .899&.825 & .022 & .922& .909& \multicolumn{1}{c|}{.684}
& .911&.863 & .045 & .897& .899 & \multicolumn{1}{c}{.706} \\
\multicolumn{1}{l|}{\textbf{LDF}\cite{wei2020label}}
& .904&.849 &.032  &.919&.904 & \multicolumn{1}{c|}{.663} 
& .911&.864 &.019  &.947&.922 & \multicolumn{1}{c|}{.667} 
& .915&.856 &.047  &.892&.890 & \multicolumn{1}{c}{.682} 
\\
\multicolumn{1}{l|}{\textbf{CTD}\cite{zhao2021complementary}}
&.906& .848  & .031 & .921 & .905 & \multicolumn{1}{c|}{.646}
&.904& .844  & .019 & .938 & .911 & \multicolumn{1}{c|}{.649}
&.919& .869  & .042 & .899 & .898 & \multicolumn{1}{c} {.669}
\\
\multicolumn{1}{l|}{\textbf{PFS}\cite{ma2021pyramidal}} 
& .911&.849 & .033 & .922 &.906 & \multicolumn{1}{c|}{.674} 
& .916&.867 & .019 & .946 &.923  & \multicolumn{1}{c|}{.688} 
& .918&.868 & .043 & .898 &.899& \multicolumn{1}{c}{.701}
\\

\multicolumn{1}{l|}{\textbf{ICON}\cite{ICON}} 
& .920&.878  &.025 &.939  &.922 & \multicolumn{1}{c|}{.636} 
& .928&.886 & .015   &.966 &.933& \multicolumn{1}{c|}{.635} 
& .933&.896& .032  &.907 &.919& \multicolumn{1}{c}{.660}
\\\toprule \hline
\multicolumn{19}{c}{\textbf{HR-SOD methods}} 
\\ \midrule \hline
\multicolumn{1}{l|}{\textbf{HRSOD-DH\cite{zeng2019towards}}}  
&.905 &.845 &.030 &.934 &.896 & \multicolumn{1}{c|}{.623}
&.899 &.817 &.026 &.955 &.876 & \multicolumn{1}{c|}{.618} 
& -  &-& - & - & - & \multicolumn{1}{c}{-} 
\\
\multicolumn{1}{l|}{\textbf{HQNet-DH\cite{tang2021disentangled}}} 
&.922 &.891 &.022 &.947 &.920& \multicolumn{1}{c|}{.690} 
&.938 &.912 &.012 &.974 &.939& \multicolumn{1}{c|}{.717}
&.927 &.880 &.039 &.899 &.900& \multicolumn{1}{c}{.715}   \\ 

\multicolumn{1}{l|}{\textbf{RMFormer-DH\cite{RMFormer}}} 
&.941 &.911 &.020 &.949 &.940 & \multicolumn{1}{c|}{.716}
&.955 &.928 &\underline{.010} &.978 &.952 & \multicolumn{1}{c|}{.717} 
&.949 &.916 &.027 &.914 &.931 & \multicolumn{1}{c}{.744}
\\
\multicolumn{1}{l|}{\textbf{RMFormer-UH\cite{RMFormer}}}
&\underline{.945} &.914 &\underline{.019} &.949 &.941 & \multicolumn{1}{c|}{.736}
&.963 &.941 &\textbf{.008} &\textbf{.982} &\underline{.958} & \multicolumn{1}{c|}{.736} 
&\textbf{.961} &\textbf{.938} &\textbf{.019} &\textbf{.921} &\textbf{.947} & \multicolumn{1}{c}{\underline{.784}}
\\ \toprule \hline
\multicolumn{19}{c}{\textbf{Our PGNet and PGNeXt}}                               \\ \midrule \hline

\multicolumn{1}{l|}{\textbf{PGNet-DH}}
& .937&.898 &.020 &.946 &.935& \multicolumn{1}{c|}{.714} 
& .950&.916 &.012 &.975 &.948 & \multicolumn{1}{c|}{.715}     
& .935&.888 &.036 &.905 &.912 & \multicolumn{1}{c}{{.736}} 
\\
\multicolumn{1}{l|}{\textbf{PGNet-UH}}
&\underline{.945}&.901 & {.020} &{.946} &{.938} & \multicolumn{1}{c|}{.727}     &.957&.929 & \underline{.010} &\underline{.979} &{.954} & \multicolumn{1}{c|}{{.730}} 
&.949&.917 & {.026} &{.916} &\underline{.935} & \multicolumn{1}{c}{{.765}}
\\
\multicolumn{1}{l|}{\textbf{PGNeXt-DH}}
&\textbf{.952} &\textbf{.929} &\textbf{.016} &\textbf{.960} &\textbf{.949} &\multicolumn{1}{c|}{\underline{.740}}
&\textbf{.966} &\textbf{.945} &\textbf{.008} &\textbf{.982} &\textbf{.960} &\multicolumn{1}{c|}{\underline{.741}}
&.950 &\underline{.920} &.026 &.913 &.934 &\multicolumn{1}{c} {.765}
\\
\multicolumn{1}{l|}{\textbf{PGNeXt-UH}}
&\underline{.945} &\underline{.920} &\underline{.019} &\underline{.953} &\underline{.944} &\multicolumn{1}{c|}{\textbf{.745}}
&\underline{.964} &\underline{.943} &\textbf{.008} &\textbf{.982} &\textbf{.960} &\multicolumn{1}{c|}{\textbf{.748}}
&\underline{.960} &\textbf{.938} &\underline{.020} &\underline{.919} &\textbf{.947} &\multicolumn{1}{c}{\textbf{.789}}
\\ \bottomrule[2pt]
\end{tabular}%

}
\end{table*}

\subsection{Results and Analysis}
\subsubsection{Quantitative Comparison on 3 HR Datasets}

We compare our proposed PGNeXt with 11 state-of-the-art existing SOD methods, of which 8 are designed for LR-SOD task, and 3 (\ie HRSOD\cite{zeng2019towards}, HQNet\cite{tang2021disentangled},RMFormer\cite{RMFormer})are specifically tailored for HR-SOD scenarios. Besides, we present the comprehensive results from these HR-SOD methods and ours under different training settings, as shown in  \cref{table:performance}.

\noindent \textbf{Comparison between LR and HR methods.} From \cref{table:performance}, we can observe that HR-SOD methods generally perform significantly better on these high-resolution datasets than LR-SOD methods. We attribute this significant improvement to two main factors, Firstly, HR-SOD methods can extract more complete structural information from high-resolution input while LR-SOD methods lose during the down-sampling process. This enables a considerable improvement in metrics reflecting overall quality \eg $E_\xi$ and $S_\alpha$. Secondly, the improvement is due to the HR-SOD methods' particular focus on the boundary details in images. It can be observed that while HQNet-DH and the top LR-SOD method ICON are close, with some even being lower in these 5 overall metrics, HQNet-DH notably excels over ICON in the mBA metric by a large margin (\ie on UHRSD-TE, \textit{0.715 \vs 0.660}). Thus, it is evident that the edges in saliency maps produced by HR-SOD methods are far superior to those obtained by LR-SOD methods. 

\noindent \textbf{Comparison among HR-SOD methods.} 
In \cref{table:performance}, we can observe that our PGNeXt outperforms all other HR-SOD methods in terms of metrics on 3 HR benchmarks. Firstly, compared to the previous state-of-the-art method, especially RMFormer \cite{RMFormer}, 
 
on the basis of other metrics to maintain advantages or comparable performance, our PGNeXt shows a marked superiority in the mBA metric, indicating a superior ability of PGNeXt to generate clearer and sharper edges in saliency maps,
thereby better satisfying the specific requirements of HR-SOD task. Additionally, \cref{table:performance} shows that our PGNeXt has substantially improved our previous conference version PGNet, proving the effectiveness of our new proposed technical contributions in this latest version. 
Last but not least, through the comparison between the performance of same model under different training settings DH and UH,

we observe an interesting point, that is, 
models trained under UH settings show minor fluctuations in overall metrics on HRSOD-TE and DAVIS-S, but with a more noticeable increase on UHRSD-TE.
We attribute it to the unique high-resolution challenge of the proposed UHRSD dataset.
Under the training setting of DH, since the distribution and characteristics of training datasets are consistent with  HRSOD-TE and DAVIS-S, the performance on these two datasets is relatively improved.
However, under the UH setting after adding the new dataset UHRSD-TR with more high-resolution details, our model not only has considerable performance on the conventional HR-SOD dataset, but also has obvious advantages on the more complex UHRSD-TE. This further verifies the challenges of our new constructed dataset, and highlights the advantages of our model in the face of high-resolution tasks.

\subsubsection{Qualitative Comparison}
\begin{figure*}[t]
    \centering
    \includegraphics[width=\textwidth]{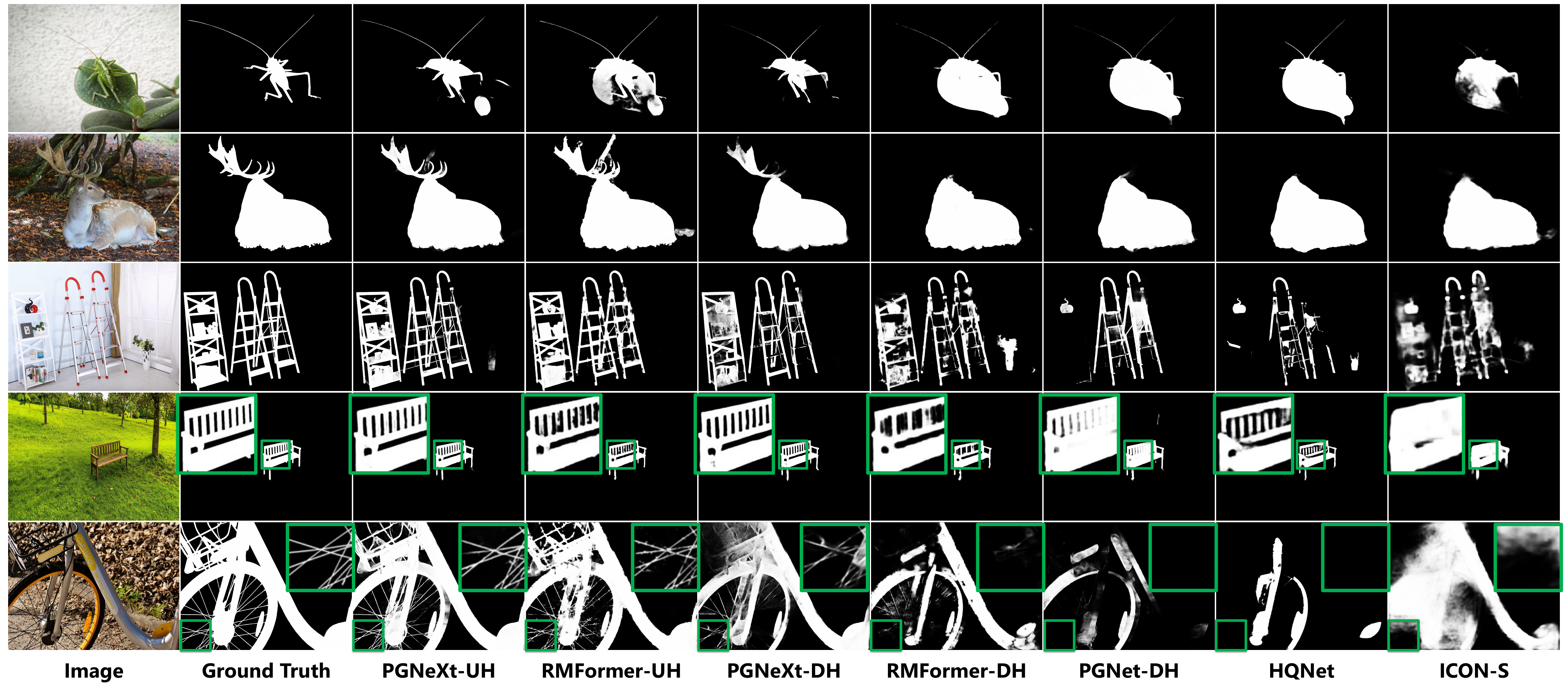}
    
    \caption{Qualitative comparisons with previous state-of-the-art methods. We show the saliency maps from both the best 2 LR-SOD methods and HR-SOD methods under different training settings for better observe the visual characteristics of different methods. We have zoomed some details and you can further zoom-in for better viewing.}
    \label{fig:visual_comparison}
\end{figure*}

In line with our previous discussion, visual results better highlight the characteristics of HR-SOD task, thus we present some saliency maps generated by different methods in \cref{fig:visual_comparison}. Firstly, it's evident that HR-SOD methods possess a definitive advantage over LR-SOD methods.
As shown in the last two columns in \cref{fig:visual_comparison}, LR-SOD methods can even not accurately locate all salient region.
Secondly, it’s noticeable that our PGNeXt outperforms other HR SOD methods in some hard regions. Specifically, as illustrated in first 2 rows in \cref{fig:visual_comparison}, our PGNeXt can capture accurate global semantics, enabling it to precisely determine whether a part belongs to potential targets such as antlers. Beyond this, PGNeXt can precisely segment extremely small structures, such as spokes in bicycle tires as shown in the last 3 rows. In contrast, other methods exhibit significantly lower segmentation quality, facing issues like inability to distinguish closely adjacent fine regions and background noise. 
Lastly, compared with DH setting, we can observe that our model trained under UH setting has more surprising performance on the fine edges and contours of objects, such as ladders, chairs and bicycles shown in the last three rows. This further proves the quantitative advantage in mBA metric as shown in \cref{table:performance}. Besides, this also emphasizes the benefits of high quality HR-SOD dataset and validates the importance of our purpose to provide a large-scale HR-SOD datasets.

\begin{table}[t]
    \centering
    \caption{Comparison of HR-SOD methods' efficiency. We conducted this assessments on a consistent platform following the original settings. FLOPs refers to the floating point operations and Params refers to the sum of trainable parameters in the models.}
    \renewcommand\arraystretch{1.3}
    \resizebox{0.95\linewidth}{!}{
    \begin{tabular}{l|cccc}
    \toprule
         Model &Resolution & FLOPs(G) &Params(M) &Speed(FPS)  \\
         \hline
         PGNeXt & $1536 \times 1536$ & 202.76 & 111.62 & 27.61 \\
         HQNet\cite{tang2021disentangled} & $1024 \times 1024$ &15.22 & 59.69 & 4.07 \\
         RMFormer\cite{RMFormer} & $1536 \times 1536$ & 1252.96 &262.36 & 7.60 \\
         PGNet\cite{PGNet} & $1024 \times 1024$ & 71.02 &72.67 &34.23 \\
    \bottomrule     
    \end{tabular}
    }
    \label{tab:speed}
\end{table}

\subsection{Benchmark on UHRSD-TE subsets}
As we described in \cref{sec:dataset}, we split the UHRSD-TE datasets to 4 subsets according to their shape complexity, aiming to evaluate the performance of different methods on high-resolution images of various levels of shape complexity. The assessment could serve as a benchmark for HR-SOD methods and inform future researches.
We evaluate 13 methods using 5 metrics and the evaluation results are shown in \cref{tab:uhrsd_te}. 
On the one hand, we can see that as the structural complexity increases, the overall trend of mBA metrics of these LR-SOD methods drop significantly. This trend implies that the previous methods are insufficient in dealing with complex boundaries on high-resolution images.
On the other hand, compared with LR-SOD method, the HR-SOD method such HQNet\cite{tang2021disentangled} and RMF\cite{RMFormer} has a great improvement in dealing with complex boundaries, especially our method has the highest mBA performance on all subsets, which also shows that the proposed grafting mechanism has excellent processing ability in high-resolution details.
This benchmark further shows that the segmentation ability at the complex boundary of HR images can be used as an important metric to measure the performance of HR models. These subsets divided according to complexity will also be released publicly.

\label{sec:subsets}

\subsection{Efficiency Comparison}
We report the efficiency comparison of several HR-SOD methods in \cref{tab:speed}. All assessments are conducted in the same environment and the input resolution is following the original settings. Compared with the HQNet\cite{tang2021disentangled}, PGNeXt has higher computation burden and parameters, but the average speed (FPS) is significantly higher. This is attributed to the fact that HQNet is a multi-stage methods, which has efficiency bottlenecks between stages, and the I/O between stages can further reduce its efficiency. And compared with the recent state-of-the-art method RMFormer\cite{RMFormer}, their floating point operations (FLOPs) and parameters are much higher, which results in a very slow inference speed of only 28.14\% of ours. In general, the high efficiency of our method is due to the fact that we asymmetrically take into account the respective characteristics of high- and low-resolution features, using lightweight structures to extract spatial information from large-size input, which keeps the inference speed fast.

\subsection{Ablation Studies}
We first provide detailed analysis of each component in PGNeXt. Furthermore, we experiment with different configuration within these components to better illustrate their effectiveness. All experiments in this section are based on the setting of \textbf{PGNeXt-UH}. And we evaluate them on HR-SOD datasets to directly demonstrate impact of each component on HR-SOD task.
\subsubsection{Ablation on components of PGNeXt}
In \cref{tab:ablation_all}, we investigate the contribution of each main components in PGNeXt including hierarchical staggered connection (HSC), window-based Cross-Model Grafting Module (wCMGM) and attention guided loss (AGL). The experiments are conducted based on the baseline composed of Swin encoder and plain decoder in traditional FPN style. Due to the component dependencies, we sequentially add HSC, wCMGM and AGL to the baseline:

\noindent \textbf{Effectiveness of HSC} We introduce extra two CNN branch to the baseline model (No. \#1 in \cref{tab:ablation_all}) and connect them in the staggered style shown in \cref{fig:3}, which is denoted as No. \#2. Comparing with No. \#1, we find that there is a significant improvement on all of the 3 datasets. In particular, the mBA on UHRSD-TE improved from 0.672 to 0.742 (10.4\% increase), which is far more significant than the rest overall structural metrics. This improvement indicates that the additional large resolution CNN branches provide rich details on top of the Swin branch's semantics, resulting in greatly clear edges in saliency maps. 

\begin{table*}[t]
    \caption{Quantitative evaluation on UHRSD-TE subsets. The best two results are highlighted in \textbf{bold} and \underline{underlined}.}
    \renewcommand\arraystretch{1.2}
    \centering
    \setlength\tabcolsep{6pt}
    \begin{tabular}{C{1cm}|L{0.6cm}|C{0.75cm}|C{0.75cm}|C{0.75cm}|C{0.75cm}|C{0.75cm}|C{0.75cm}|C{0.65cm}|C{0.9cm}|C{0.9cm}|C{0.85cm}|C{0.85cm}|C{0.75cm}|C{0.75cm}}
    \toprule 
    Dataset & Metric &\scriptsize F3Net\cite{wei2020f3net}&\scriptsize ITSD\cite{zhou2020interactive}&\scriptsize LDF\cite{wei2020label}&\scriptsize CTDNet\cite{zhao2021complementary}&\scriptsize PFSNet\cite{ma2021pyramidal}&\scriptsize ICON-S\cite{ICON}&\scriptsize HQNet\cite{tang2021disentangled}& {\scriptsize RMF-DH\cite{RMFormer}}&{\scriptsize RMF-UH\cite{RMFormer}} &\scriptsize PGNet-DH\cite{PGNet}&\scriptsize PGNet-UH\cite{PGNet}&\scriptsize PGNeXt-DH&\scriptsize PGNeXt-UH \\
    \hline
    \multirow{6}{\linewidth}{\centering UHRSD-TE1}
&$\mathcal{M}$  &0.056  &0.052  &0.061  &0.050  &0.050  &0.033  &0.049  &0.031  &\textbf{0.020}  &0.043  &0.027  &0.030  &\underline{0.023}\\
&$F_\beta^{w}$  &0.879  &0.889  &0.869  &0.891  &0.893  &0.926  &0.897  &0.933  &\textbf{0.951}  &0.903  &0.932  &\underline{0.936}  &\textbf{0.951}\\
&$S_\alpha$   &0.895  &0.903  &0.888  &0.903  &0.906  &0.930  &0.903  &0.934  &\textbf{0.951}  &0.914  &0.939  &0.938  &\underline{0.950}\\
&$E_\xi$   &0.868  &0.881  &0.873  &0.870  &0.878  &0.881  &0.872  &\textbf{0.889}  &0.880  &0.878  &\underline{0.883}  &\underline{0.883}  &0.880\\
&$\mathrm{mBA}$  &0.696  &0.716  &0.691  &0.678  &0.708  &0.671  &0.719  &0.747  &\underline{0.780}  &0.739  &0.767  &0.765  &\textbf{0.786}\\
    \hline
    \multirow{6}{\linewidth}{\centering UHRSD-TE2}
&$\mathcal{M}$  &0.042  &0.042  &0.043  &0.041  &0.039  &0.031  &0.038  &0.026  &\underline{0.022}  &0.032  &0.027  &0.024  &\textbf{0.021}\\
&$F_\beta^{w}$  &0.858  &0.866  &0.855  &0.871  &0.873  &0.901  &0.886  &0.917  &\underline{0.933}  &0.901  &0.913  &0.922  &\textbf{0.938}\\
&$S_\alpha$   &0.896  &0.902  &0.893  &0.899  &0.903  &0.923  &0.905  &0.933  &\underline{0.944}  &0.921  &0.933  &0.937  &\textbf{0.947}\\
&$E_\xi$   &0.888  &0.904  &0.891  &0.901  &0.905  &0.915  &0.907  &0.918  &0.924  &0.913  &\underline{0.926}  &0.919  &\textbf{0.930}\\
&$\mathrm{mBA}$  &0.687  &0.709  &0.685  &0.669  &0.703  &0.661  &0.719  &0.746  &\underline{0.784}  &0.742  &0.767  &0.765  &\textbf{0.792}\\

    \hline
    \multirow{6}{\linewidth}{\centering UHRSD-TE3}
&$\mathcal{M}$  &0.038  &0.037  &0.040  &0.034  &0.035  &0.030  &0.035  &0.023  &\underline{0.016}  &0.033  &0.021  &0.020  &\textbf{0.015}\\
&$F_\beta^{w}$  &0.871  &0.875  &0.870  &0.885  &0.880  &0.895  &0.884  &0.925  &\underline{0.947}  &0.888  &0.932  &0.936  &\textbf{0.949}\\
&$S_\alpha$   &0.904  &0.911  &0.900  &0.911  &0.911  &0.921  &0.905  &0.938  &0.954  &0.913  &\underline{0.947}  &0.945  &\textbf{0.955}\\
&$E_\xi$   &0.918  &0.922  &0.914  &0.932  &0.922  &0.931  &0.922  &0.938  &\textbf{0.949}  &0.924  &0.943  &0.941  &\underline{0.947}\\
&$\mathrm{mBA}$  &0.687  &0.710  &0.685  &0.671  &0.707  &0.657  &0.719  &0.749  &\underline{0.793}  &0.740  &0.775  &0.774  &\textbf{0.798}\\
    \hline
    \multirow{6}{\linewidth}{\centering UHRSD-TE4}
&$\mathcal{M}$  &0.047  &0.047  &0.044  &0.043  &0.046  &0.034  &0.035  &\underline{0.029}  &\textbf{0.020}  &0.036  &\underline{0.029}  &\underline{0.029}  &\textbf{0.020}\\
&$F_\beta^{w}$  &0.816  &0.823  &0.833  &0.831  &0.826  &0.860  &0.861  &0.891  &\textbf{0.922}  &0.860  &0.891  &0.891  &\underline{0.919}\\
&$S_\alpha$   &0.872  &0.880  &0.881  &0.881  &0.876  &0.902  &0.894  &0.918  &\underline{0.938}  &0.898  &0.921  &0.918  &\textbf{0.939}\\
&$E_\xi$   &0.881  &0.883  &0.893  &0.893  &0.887  &0.902  &0.904  &0.912  &\textbf{0.929}  &0.904  &0.912  &0.912  &\underline{0.921}\\
&$\mathrm{mBA}$  &0.668  &0.689  &0.671  &0.657  &0.686  &0.652  &0.710  &0.736  &\underline{0.779}  &0.724  &0.752  &0.756  &\textbf{0.782}\\
    \bottomrule
    
    \end{tabular}
    \label{tab:uhrsd_te}
    
\end{table*}

\noindent \textbf{Effectiveness of wCMGM.} 
Compared with No. \#2, we replace the plain fusion strategy with our proposed wCMGM module at grafting points between branches, which is shown as No. \#3. We can observe that the metrics on 3 datasets are further improved substantially. Compared to the performance gain from No. \#1 to No. \#2, this improvement is mainly in the overall metrics, \eg increase from 0.912 to 0.914 \textit{vs.} 0.914 to 0.920 in terms of $E_\xi$ on UHRSD-TE. Besides, the mBA metric exhibits a slight increase, which could partially be attributed to enhanced accuracy in locating salient regions. We believe the boost is due to the fact that wCMGM makes the auxiliary CNN branches not only provide details in large resolution, but also graft the context from global branch to local branches and make the heterogeneous feature complementary from each other, which ultimately leads to a more accurate saliency identification.

\begin{figure}[t]
    \centering
    \includegraphics[width=\linewidth]{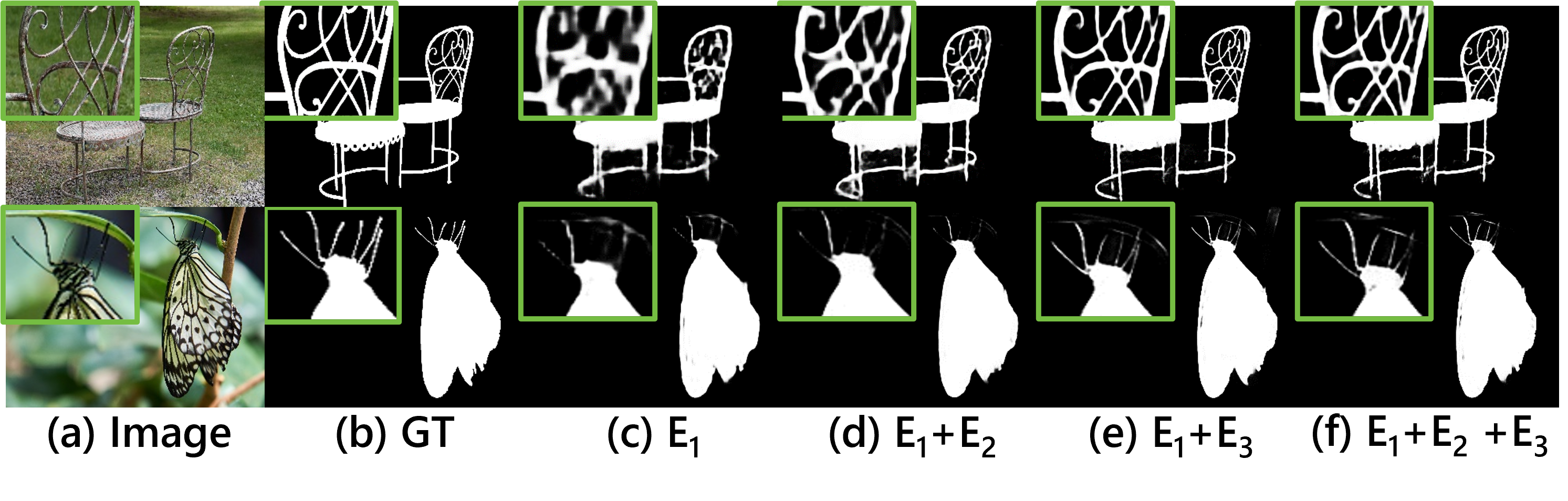}
    \caption{Visualization comparisons for saliency maps among different combinations of branches in hierarchical staggered connection architecture. Best viewed by zoom-in.}
    \label{fig:ab}
\end{figure}
\noindent \textbf{Effectiveness of AGL.} 
As shown in last row in \cref{tab:ablation_all}, the adoption of AGL brings performance growth in terms of all metrics. In specific, the increase is minor in terms of $S_\alpha$ and $E_\xi$, which mainly reflects the accuracy of global structure of salient objects. And the main growth is in $\mathcal{M}$ and $mBA$, indicating AGL allows the model to concentrate on edge regions of salient objects, making the overall pixel-level error rate lower. 

\begin{table*}[t]
    \caption{Ablation analysis on the effectiveness of each component in PGNeXt. FPN denotes the baseline model of PGNeXt which adopts the preliminary FPN consisting of Swin encoder and simple decoder. HSC denotes the hierarchical staggered connection. The best results are highlighted in \textbf{bold}.}
    \LARGE
    \renewcommand\arraystretch{1.3}
    \centering
    \setlength\tabcolsep{6pt}
    \resizebox{0.95\textwidth}{!}{%
    \begin{tabular}{c|cccc|ccccc|ccccc|ccccc}
    \toprule
    \multirow{2}{*}{\centering \textbf{No.}} & \multicolumn{4}{c|}{Component} & \multicolumn{5}{c|}{UHRSD-TE} & \multicolumn{5}{c|}{HRSOD-TE} & \multicolumn{5}{c}{DAVIS-S} \\
    \cline{2-20}
    &FPN &HSC &wCMGM &AGL &{$\mathcal{M}$}&{$S_\alpha$}& $E_\xi$&$F_\beta^{M}$&\multicolumn{1}{c|}{mBA}&{ $\mathcal{M}$}&{$S_\alpha$}& $E_\xi$&$F_\beta^{M}$&\multicolumn{1}{c|}{mBA}&{ $\mathcal{M}$}&{ $S_\alpha$}& $E_\xi$& $F_\beta^{M}$&\multicolumn{1}{c}{mBA}\\
    \hline
         \#1 &\checkmark &           &           &           
           & 0.026 & 0.933 & 0.912 & 0.945 & 0.672 
           & 0.025 & 0.927 & 0.937 & 0.926 & 0.644
           & 0.013 & 0.944 & 0.973 & 0.940 & 0.645\\
         \#2 &\checkmark &\checkmark &           &
           & 0.023 & 0.939 & 0.914 & 0.954 & 0.742
           & 0.022 & 0.935 & 0.946 & 0.934 & 0.702
           & 0.011 & 0.947 & 0.976 & 0.951 & 0.699\\
         \#3 &\checkmark &\checkmark &\checkmark &
           & 0.022 & 0.944 & 0.914 & 0.956 & 0.776
           & 0.021 & 0.940 & \textbf{0.949} & 0.941 & 0.731
           & 0.009 & 0.957 & 0.980 & 0.962 & 0.726\\
         \#4 &\checkmark &\checkmark &\checkmark &\checkmark 
           & \textbf{0.019} & \textbf{0.947} & \textbf{0.921} & \textbf{0.961} & \textbf{0.784}
           & \textbf{0.019} & \textbf{0.941} & \textbf{0.949} & \textbf{0.945} & \textbf{0.736}
           & \textbf{0.008} & \textbf{0.958} & \textbf{0.982} & \textbf{0.963} & \textbf{0.736}\\
         \bottomrule
    \end{tabular}
    }
    
    \label{tab:ablation_all}
\end{table*}

\subsubsection{Ablation on hierarchical staggered connection} 
We delve into further exploration on the effect of hierarchical staggered connection. \cref{tab:ablation_layer} shows  different combinations of feature pyramids, and the labels $E_i$ of encoders can correspond to \cref{fig:3}. No. \ding{72}1 denotes the baseline model which only adopt the Swin encoder to construct the feature pyramid. In variants No. \ding{72}2 and No. \ding{72}3, we add the corresponding encoder branch to No. \ding{72}1, respectively, to aggregate features from branches of different resolution with our proposed staggered connection strategy to form a higher feature pyramid. We find that adding the $E_2$ branch gives a performance boost, but it is relatively limited. However, adding $E_3$ encoder the mBA metrics are significantly improved as shown in 3-th row in \cref{tab:ablation_layer}. Analyzing this result, we conclude that $E_2$ branch has a small resolution increase compared to $E_1$, providing limited detail and not enough to improve the edge quality. On the other hand, $E_3$ branch is 4 times larger than $E_1$ branch in terms of the feature map size, and the network can learn more details from the large size feature maps to generate saliency maps with high edge quality. This is evidenced by the visual comparison in \cref{fig:ab}. 
Besides, we note that while the edge quality is greatly improved, the $E_\xi$ metric decrease relative to No. \ding{72}2. We believe this is due to the fact that while $E_3$ branch brings rich details, it also brings some error as well as background noise, because the relative receptive field is limited. Lastly, the variant No. \ding{72}4 is the hierarchical staggered connection we used in PGNeXt.  In \cref{tab:ablation_layer}, the $\mathcal{M}$ and $E_\xi$ in the last row are increased a lot compared to the other rows. This is due to the hierarchical staggered connection that gradually aggregates features from low-resolution branch to higher-resolution branch. This process narrows the semantic gap among the three branches and provides a progressive fusion effect, enabling more accurate identification of salient regions. Besides, we find that compared to \ding{72}3, the mBA decreases, and we speculate that this may result from the incorporation of the relative coarse $E_2$ causing the fusion of the edge region to be confused, which also implies that we need further grafting strategy instead of the plain feature fusion.

\begin{table}[t]
    \caption{Ablation analysis on the number of branches. $E_i$ denotes the involvement of the $i$-th encoder in the pyramid feature grafting process. The best results are highlighted in \textbf{bold}.}
    \Large
    \renewcommand\arraystretch{1.27}
    \centering
    \setlength\tabcolsep{6pt}
    \resizebox{0.9\linewidth}{!}{%
    \begin{tabular}{c|ccc|ccc|ccc}
    \toprule
    \multirow{2}{*}{\centering \textbf{No.}} & \multicolumn{3}{c|}{Branch} & \multicolumn{3}{c|}{UHRSD-TE} & \multicolumn{3}{c}{HRSOD-TE} \\
    \cline{2-10}
    & ${E_1}$ &${E_2}$ &${E_3}$& $\mathcal{M}$ & $E_\xi$ &\multicolumn{1}{c|}{mBA} & $\mathcal{M}$ & $E_\xi$ &\multicolumn{1}{c}{mBA}\\
    \hline
    \ding{72}1 &\checkmark &           &           
    &0.026& 0.912& 0.672& 0.025& 0.937& 0.644 \\
    \ding{72}2 &\checkmark &\checkmark &           
    &0.026&0.913&0.673&0.025&0.944&0.653 \\
    \ding{72}3 &\checkmark & &\checkmark 
    &0.024&0.911&\textbf{0.751}&0.024&0.941&\textbf{0.707} \\
    \ding{72}4 &\checkmark &\checkmark &\checkmark 
    &\textbf{0.023}&\textbf{0.914}&{0.742}&\textbf{0.022}&\textbf{0.946}&{0.702} \\
    \bottomrule
    \end{tabular}
    }
    
    \label{tab:ablation_layer}
\end{table}

\subsubsection{Ablation on wCMGM}
We provide the effect of different readjustment configuration in wCMGM on the performance in \cref{tab:ablation_alpha}. In specific, we vary the $\alpha$ in \cref{eq:alpha} to adjust the amplification effect for the cross attention maps. The first row in \cref{tab:ablation_alpha} is the result without readjusting strategy in wCMGM, and metrics are lower than results in the rest rows. It indicate that the readjustment in wCMGM is necessary for improving the performance. And in the comparison among \ding{83}2, \ding{83}3 and \ding{83}4, we can observe that the performance is relatively close, and overal the best performance is achieved when $\alpha=2$.

\begin{table}[t]
    \caption{Ablation analysis on readjustment strategy in wCMGM. The best results are highlighted in \textbf{bold}.}
    \Large
    \renewcommand\arraystretch{1.5}
    \centering
    \setlength\tabcolsep{6pt}
    \resizebox{\linewidth}{!}{%
    \begin{tabular}{c|c|cccc|cccc}
    \toprule
    \multirow{2}{*}{\centering \textbf{No.}} & \multirow{2}{*}{Readjust} & \multicolumn{4}{c|}{UHRSD-TE} & \multicolumn{4}{c}{HRSOD-TE} \\
    \cline{3-10}
    && $\mathcal{M}$ & $S_\alpha $ & $E_\xi$ &\multicolumn{1}{c|}{mBA} & $\mathcal{M}$ & $S_\alpha$ & $E_\xi $ &\multicolumn{1}{c}{mBA}\\
    \hline
    \ding{83}1 &\textit{w/o} 
    &0.022 &0.944&0.914& 0.776& 0.021& 0.940& \textbf{0.949}& 0.731 \\
    \ding{83}2 &$\alpha=1$            
    &0.021&0.945&0.915&0.776&0.020&{0.941}& \textbf{0.949} &0.729 \\
    \ding{83}3 &$\alpha=2$ 
    &\textbf{0.019}&\textbf{0.947}&\textbf{0.921}&\textbf{0.784}&\textbf{0.019}&{0.941} &\textbf{0.949} &\textbf{0.736} \\
    \ding{83}4 &$\alpha=3$ 
    &0.021&0.946&0.918&0.778&0.021&\textbf{0.942}&\textbf{0.949}&0.732 \\
    \bottomrule
    \end{tabular}
    }
    
    \label{tab:ablation_alpha}
\end{table}

\section{Generalization To Camouflaged Object Detection}
In this section, to demonstrate the generalization ability, we apply our PGNeXt to the closely related challenging task of camouflaged object detection (COD)\cite{COD}, which aims to locate and segment objects blended into the background. We chose the COD task not only because it is a binary segmentation task like SOD, but also for two main reasons: 1) COD datasets typically have higher resolution to distinguish small camouflaged objects and show inconspicuous edge clues; 2) Existing researches \cite{COD,HIT} indicate that the performance of COD methods is greatly influenced by input resolution. Hence, we argue that the application to COD task can reflect the PGNeXt's comprehensive capability in handling high-resolution images.
\subsection{Implementation Details}

To verify the generalization ability of PGNeXt, we do not modify any modules when applying on COD task. For a fair comparison, we follow the training setting of recent COD methods \cite{COD,HIT,FRI} using the combination of COD10K and CAMO datasets for training process. And we evaluate our models on 3 widely-used COD benchmarks, \ie COD10K\cite{COD}, CAMO\cite{CAMO} and NC4K\cite{Rank-Net}. 

We set the maximum learning rate to 0.06, with the learning rate for Swin-B at 0.02 of the maximum learning rate. The size of mini-batch is set to 32 and the whole training process takes 56 epochs. The remaining training settings are similar to those used in HR-SOD task.

\subsection{Quantitative Comparison}

In \cref{tab:cod_quantity} we show the quantity results of several state-of-the-art COD methods and our PGNeXt on 3 widely-used benchmarks.  It is noteworthy that HitNet\cite{HitNet} ,which has impressive performance in  \cref{tab:cod_quantity}, is also designed with the objective of detecting camouflaged objects in high-resolution. As can be seen, without bells and whistles our PGNeXt surpasses most of existing COD methods by a large margin. We argue that the high-resolution design of PGNeXt effectively aids in detecting and segmenting small targets of structures that are challenging to discern at lower resolution, thereby contributing to its strong performance in COD task. The success of PGNeXt applying on COD task amply validates its generalization ability in similar binary segmentation tasks.

\begin{figure*}[t]
    \includegraphics[width=0.9\textwidth]{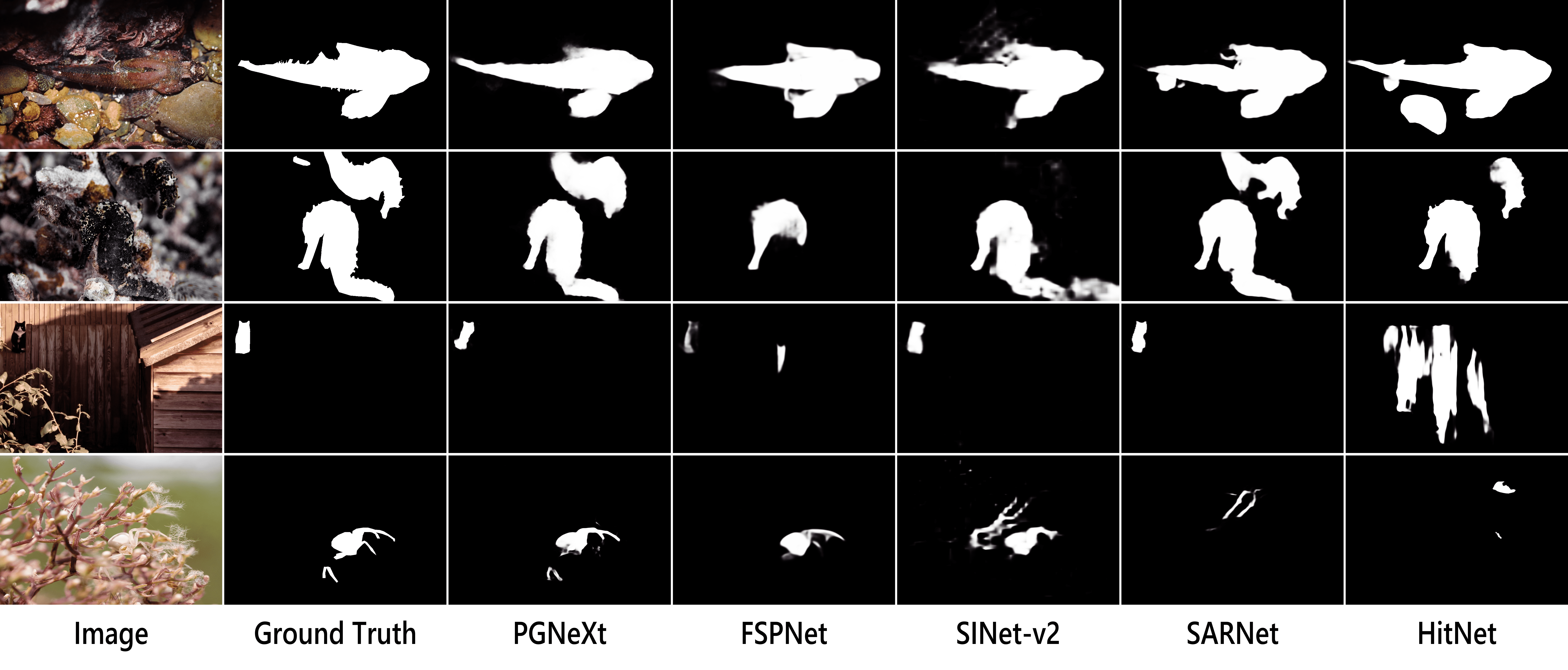}
    \centering
    \caption{Visual comparison with state-of-the-art COD methods. Best viewed by zoom-in.}
    \label{cod_vis}
\end{figure*}

\subsection{Qualitative Comparison}
We show visual comparison between PGNeXt and existing state-of-the-art COD methods including FSPNet\cite{FSPNet}, SINet-V2\cite{COD}, SARNet\cite{SARNet} and HitNet\cite{HitNet}. As can be seen in \cref{cod_vis}, our PGNeXt is not only capable of accurately detecting multi-scale concealed objects, but it can also precisely segment the blurred boundaries of them. As shown in first two rows in \cref{cod_vis}, the global information captured by the transformer branch enables PGNeXt to precisely identify camouflaged objects, ensuring neither omissions nor redundancy. From rows 3 and 4 we can clearly see that PGNeXt is capable of segmenting extremely small targets. While HitNet, another high-resolution method tailed for COD task, preserves sharp edges like PGNeXt, but it suffers from serious misidentification of camouflaged targets. Both quantitative and qualitative results demonstrate the potential of PGNeXt generalizing on other low-level segmentation tasks.

\begin{table}[t]
  {
  \small
  \caption{Quantitative comparison of our PGNeXt with state-of-the-art COD methods on three benchmark datasets including CAMO, COD10K and NC4K. The best results of each metric are in \textbf{bold}}
  \renewcommand\arraystretch{1.3}
  \label{tab:comp}
  
  \begin{tabular}{L{1.8cm}|C{0.23cm}C{0.23cm}C{0.23cm}|C{0.23cm}C{0.23cm}C{0.23cm}|C{0.23cm}C{0.23cm}C{0.23cm}}
    \toprule
    \multirow{2}{*}{\centering \textbf{Method}}  & \multicolumn{3}{c|}{\textbf{CAMO}} & \multicolumn{3}{c|}{\textbf{COD10K}} & \multicolumn{3}{c}{\textbf{NC4K}}\\
    \cline{2-10}
    &{\scriptsize $\mathcal{M}$}&{\scriptsize $S_\alpha$}&\multicolumn{1}{c|}{\scriptsize $F_\beta^{M}$}&{\scriptsize $\mathcal{M}$}&{\scriptsize $S_\alpha$}&\multicolumn{1}{c|}{\scriptsize $F_\beta^{M}$}&{\scriptsize $\mathcal{M}$}&{\scriptsize $S_\alpha$}&\multicolumn{1}{c}{\scriptsize $F_\beta^{M}$}\\
    \midrule
    SINet{\footnotesize~\cite{SINet}}
    &.100&.751&\multicolumn{1}{c|}{.675}
    &.051&.771&\multicolumn{1}{c|}{.634} 
    &.058&.808&\multicolumn{1}{c}{.769}\\
    Rank-Net{\footnotesize~\cite{Rank-Net}}
    &.080&.787&\multicolumn{1}{c|}{.744} 
    &.037&.804&\multicolumn{1}{c|}{.715} 
    &.048&.840&\multicolumn{1}{c}{.804}\\ 
    UGTR{\footnotesize~\cite{UGTR}}
    &.086&.784&\multicolumn{1}{c|}{.735} 
    &.036&.817&\multicolumn{1}{c|}{.712} 
    &.052&.839&\multicolumn{1}{c}{.787}\\
    BSA-Net{\footnotesize~\cite{BSA-Net}}
    &.079&.794&\multicolumn{1}{c|}{.763} 
    &.034&.818&\multicolumn{1}{c|}{.738} 
    &.048&.841&\multicolumn{1}{c}{.808}\\
    SegMaR{\footnotesize~\cite{SegMaR}}
    &.071&.815&\multicolumn{1}{c|}{.795} 
    &.034&.833&\multicolumn{1}{c|}{.757} 
    &.046&.841&\multicolumn{1}{c}{.821}\\
    ZoomNet{\footnotesize~\cite{ZoomNet}}
    &.066&.820&\multicolumn{1}{c|}{.794} 
    &.029&.838&\multicolumn{1}{c|}{.766} 
    &.043&.853&\multicolumn{1}{c}{.818}\\
    SINet-v2{\footnotesize~\cite{COD}}
    &.070&.820&\multicolumn{1}{c|}{.782} 
    &.037&.815&\multicolumn{1}{c|}{.718} 
    &.048&.847&\multicolumn{1}{c}{.805}\\
    FAPNet{\footnotesize~\cite{FAPNet}}
    &.076&.815&\multicolumn{1}{c|}{.776} 
    &.036&.822&\multicolumn{1}{c|}{.731} 
    &.047&.851&\multicolumn{1}{c}{.810}\\
    FEDER{\footnotesize~\cite{FEDER}}
    &.071&.802&\multicolumn{1}{c|}{.781} 
    &.032&.822&\multicolumn{1}{c|}{.751} 
    &.044&.847&\multicolumn{1}{c}{.824}\\
    FSPNet{\footnotesize~\cite{FSPNet}}
    &.050&.856&\multicolumn{1}{c|}{.831} 
    &.026&.851&\multicolumn{1}{c|}{.769} 
    &.035&.879&\multicolumn{1}{c}{.843}\\
    SARNet{\footnotesize~\cite{SARNet}}
    &\textbf{.047}&\textbf{.868}&\multicolumn{1}{c|}{.850} 
    &.024&.864&\multicolumn{1}{c|}{.800} 
    &.032&\textbf{.886}&\multicolumn{1}{c}{.863}\\
    HitNet{\footnotesize~\cite{HitNet}}
    &.055&.849&\multicolumn{1}{c|}{.831} 
    &.023&\textbf{.871}&\multicolumn{1}{c|}{.823} 
    &.037&.875&\multicolumn{1}{c}{.854}\\
    PGNeXt
    &\textbf{.047} &.866 &\multicolumn{1}{c|}{\textbf{.872}} 
    &\textbf{.023} &.861 &\multicolumn{1}{c|}{\textbf{.833}} 
    &\textbf{.031} &\textbf{.886} &\multicolumn{1}{c}{\textbf{.885}}  \\
    
  \bottomrule
  \end{tabular}
  \label{tab:cod_quantity}
  }
\end{table}

\section{Conclusion}
This paper presented an advanced study on the high-resolution salient object detection (HR-SOD) from both dataset and network framework perspectives. 
Specifically, we comprehensively analyzed the challenges of HR-SOD and the disadvantages of existing HR-SOD methods, and proposed the pyramid grafting mechanism. Furthermore, we designed the window-based Cross-Model Grafting Module (wCMGM) and attention guided loss (AGL), based on which we constructed the PGNeXt.
Considering the lack of available HR-SOD datasets, we elaborately organize and annotate ultra-high-resolution saliency detection (UHRSD) dataset. The high quantity and quality of UHRSD will provide more opportunity for future research in HR-SOD field.
We have conducted a HR-SOD benchmark for cutting-edge SOD methods on both existing HR-SOD datasets and our newly proposed UHRSD. Extensive experiments demonstrate that our PGNeXt outperforms state-of-the-art methods. Besides, the generalization capabilities shown by PGNeXt represent a great potential for many real-world applications.
\bibliographystyle{IEEEtran}
\bibliography{egbib}

\newpage

\vfill

\end{document}